\pdfoutput=1
\documentclass{article}

\usepackage[preprint]{corl_2026} %
\usepackage{times}

\usepackage[numbers]{natbib}
\usepackage{multicol}
\usepackage{amssymb}
\usepackage{amsmath}
\usepackage[utf8]{inputenc}   %
\usepackage[T1]{fontenc}      %
\usepackage[polish,english]{babel}
\usepackage{algorithm}
\usepackage{subcaption}
\usepackage{algorithmicx,algpseudocode}
\usepackage{graphicx}
\usepackage{float}
\usepackage{wrapfig}
\usepackage{booktabs}
\usepackage{siunitx, threeparttable, tabularx}
\usepackage{multirow}

\usepackage{adjustbox}
\usepackage{algorithmicx}
\usepackage{algpseudocode}
\newif\ifappendix
\appendixtrue  %

\newcommand{\suppref}[1]{%
  \ifappendix
    Appendix~\ref{#1}%
  \else
    the supplementary material%
  \fi}
\title{Task-Space Imitation Guidance for Efficient Reinforcement Learning}

\author{
  Salar Asayesh, Hossein Darani, Todd Cao, Evgeny Andriash, Mani Ranjbar\\
  Sanctuary AI,
  Canada\\
  \texttt{\{salar.asayesh,hossein.darani,todd.cao,} \\
  \texttt{evgeny.andriash,mani.ranjbar\}@sanctuary.ai} \\
}

\begin{document}
\maketitle

\begin{abstract}
  We introduce Task-Space Imitation Guidance for Efficient Reinforcement Learning (TIGER), a reward-construction and pretraining framework for sparse-reward tabletop robotic manipulation.
  TIGER treats an action-chunked imitation policy not as an executable controller or action prior, but as a local task-space progress estimator:
  predicted action chunks are converted, using controller-aware action-to-motion mapping, into short-horizon end-effector references,
  and the RL agent receives dense progress rewards toward these references while the sparse environment reward remains the dominant objective. During pretraining,
  TIGER uses imitation-guided look-ahead signals to relax conservative value penalties for actions predicted to make task-space progress,
  reducing off-manifold exploration during early online RL. Across simulation and real-robot experiments,
  TIGER improves early sample efficiency and reduces measured safety violations while matching or improving final success rates relative to prior RL and IL–RL baselines on the evaluated tasks.
  Videos and an overview of this work are available at \url{https://sanctuary.ai/news/tiger/}.

\end{abstract}

\section{Introduction}
\label{sec:intro}

Robotic imitation learning (IL) has become a central paradigm for learning
manipulation skills from demonstrations, with action chunking, diffusion
policies, and generalist robot policies showing strong results on diverse
tasks~\citep{act, dp1, pi0}.
By learning directly from demonstrations, these methods avoid manually
engineered dense rewards and accurate task dynamics models.
However, deployment-oriented manipulation exposes limitations that are less
visible in benchmark settings: policies must maintain high success under
variation while also satisfying reliability and cycle-time constraints.
Achieving this with pure IL typically requires broad, high-quality
demonstration coverage, making physical data collection costly through
teleoperation or careful curation.
More fundamentally, imitation objectives optimize agreement with demonstrated
actions rather than task success, robustness, or execution time.
Thus, IL provides a strong behavioral prior, but not a mechanism for autonomous
task optimization or systematic improvement beyond demonstrations~\citep{parl,
kim2025finetuningvisionlanguageactionmodelsoptimizing, ankile2024juicer}.

Reinforcement Learning (RL) can improve beyond imitation by directly optimizing
task success through interaction.
Unlike IL, RL can collect new experience and discover behaviors that are more
reliable or efficient than the demonstrations.
Discounted objectives also favor receiving task rewards earlier, making RL
appealing when cycle time matters.
However, direct robot RL remains difficult: sparse or binary rewards provide
little feedback during early exploration, while unstructured exploration can
cause unsafe motions, workspace violations, object drops, or frequent manual
resets.
Thus, RL provides a principled mechanism for task optimization, but requires
structured guidance to become sample-efficient and practical for real-world
manipulation.

Prior IL--RL methods improve sample efficiency by using demonstrations for
initialization, replay-buffer augmentation, auxiliary losses, or action
proposals~\citep{serl, rlpd, ibrl}.
However, in sparse-reward settings these approaches can still rely on unreliable
early critics to balance imitation and exploration.
Dense trajectory-alignment rewards offer another source of feedback~\citep{
rank2reward, LaNE}, but typically require ground-truth state trajectories,
image-level alignment, or task-specific supervision.
This motivates extracting dense guidance from imitation itself: demonstrations
naturally specify task-relevant motion, while hand-engineered rewards require
detailed state, models, or task-specific design.
Rather than using imitation only as an action prior, such guidance can provide
task-aligned feedback before sparse rewards or accurate value estimates are
available.

We propose \textbf{Task-Space Imitation Guidance for Efficient Reinforcement
Learning (TIGER)}, which turns imitation policies into dense task-space rewards
and imitation-guided critic pretraining for sparse-reward robotic manipulation.
TIGER treats an imitation policy not as an executable controller or action
prior, but as a local task-space progress estimator: given the current
observation, the policy predicts an action chunk that is mapped to a short-horizon
end-effector reference trajectory.
The RL agent receives dense reward for making progress toward this reference,
while the sparse environment reward remains the dominant task objective.
TIGER also uses imitation-guided look-ahead signals to modulate conservative
critic pretraining, reducing early off-manifold exploration.
Because guidance enters through rewards and pretraining, TIGER can
combine with different off-policy RL algorithms.
TIGER targets tasks where short-horizon end-effector motion is informative for
progress, assuming reliable end-effector pose estimates and an approximate
action-to-end-effector mapping.

We evaluate TIGER in simulation and on real robots across sparse-reward
manipulation tasks with randomized initial conditions, including grasping,
opening, and insertion.
Across settings, TIGER improves early sample efficiency, reduces measured safety
violations, and achieves competitive or higher final success rates than prior RL
and IL--RL baselines.

Our contributions are:
(i) an imitation-guided reward construction method that maps predicted action
chunks into short-horizon task-space reference motion and dense progress rewards
for sparse-reward manipulation;
(ii) an offline pretraining objective that uses imitation-guided
look-ahead signals to stabilize value learning before online interaction; and
(iii) simulated and real-robot evidence that TIGER improves early sample
efficiency and reduces measured safety violations while maintaining competitive
final success rates.
\section{Related Work}
\label{sec:related_works}

A related line of research studies learning from fixed datasets---offline reinforcement learning---as formalized by \citet{levine2020offline}, where off-policy methods must mitigate value overestimation from out-of-distribution (OOD) actions. \citet{cql} introduce pessimistic regularization that lowers the value of unsupported actions. \citet{iql} constrain policy learning to the dataset support, and~\citet{calql} calibrate value estimates to stabilize online fine-tuning. Building on these advances, several approaches combine offline pretraining with online fine-tuning~\citep{calql,awac,haldar2022watch}, achieving substantial gains in sample efficiency, though most assume access to dense rewards.

A complementary line of work uses demonstrations to address sparse or binary rewards in contact-rich manipulation.
Replay mixing with demonstrations~\citep{rlpd,vecerik2018leveragingdemonstrationsdeepreinforcement} can accelerate
learning, but becomes less effective as task horizons grow. Other methods regularize the policy toward the 
demonstrated action distribution~\citep{NEURIPS2021_eecca5b6,zhu2018dexterousmanipulationdeepreinforcement}.
GAIL~\citep{GAIL} and AIRL~\citep{fu2018airl} instead learn imitation objectives through adversarial
discriminator or reward optimization.
\citet{LI_ICLR2026_927eae0f} learns a state-only goal-proximity reward from constrained demonstrations
and extends it to online states 
through confidence-guided interpolation. TIGER uses a fixed imitation policy to provide task-space
guidance without online discriminator or reward learning.

Recent advances show that end-to-end reinforcement learning on physical robots is feasible within practical time budgets. For instance, \citet{serl}, building on \citet{rlpd}, introduce a unified framework where careful system design enables scalable real-world training from scratch, though many approaches restrict operational space and focus on short-horizon tasks. In contrast, \citet{ibrl} train an imitation policy from few demonstrations to guide exploration and bootstrap value estimation, achieving state-of-the-art efficiency but lacking a principled early-stage balance between imitation and exploration, which can slow or destabilize convergence.

More recently, dense surrogate rewards from learned perceptual or task-space representations have guided exploration in sparse-reward settings. For example, \citet{LaNE} employ a pretrained DINOv2 encoder to construct an embedding space for forward prediction, encouraging exploration near demonstration trajectories and enabling efficient real-robot learning, albeit under constrained randomization and increased computational cost.
ReWiND~\citep{rewind}, Robometer~\citep{liang2026robometer}, and RoboReward~\citep{RoboReward}
instead estimate task progress in learned visual or vision-language spaces.
TIGER converts imitation-predicted action chunks into short-horizon task-space progress signals.
\section{Preliminaries}
\label{sec:preliminaries}

We consider image-based robotic manipulation with sparse task rewards
$r^{\mathrm{env}}_t$. At time $t$, the agent observes $s_t$, consisting of
RGB images and robot proprioception, and executes action $a_t$. We train a
policy $\pi_\theta(a_t|s_t)$ and critic $Q_\theta(s_t,a_t)$ using
off-policy RL. For task-space guidance, we also use the robot end-effector pose
$x_t=(p_t,q_t)$, with position $p_t \in \mathbb{R}^3$ and unit quaternion
$q_t \in \mathbb{H}$.

We assume access to a pretrained imitation policy
$\pi^{\mathrm{IL}}_\psi$, trained by behavior cloning from demonstrations.
In our experiments, this policy is implemented as a flow-matching behavior-cloning model that predicts short action chunks
$\mathbf{a}_{t:t+T_a-1} = [a_t,\ldots,a_{t+T_a-1}]$ conditioned on the current
observation, optionally with a goal or language instruction.

\section{Method}
\label{method}
The method consists of three components: controller-aware trajectory
mapping, imitation-guided reward and observation design, and offline
pretraining with a task-aware conservative regularizer.

\subsection{Controller-Aware Trajectory Mapping}
\label{subsec:controller_aware_trajectory_mapping}

We convert each imitation-predicted action chunk into an execution-aligned EEF
trajectory using a controller-aware mapping. This avoids directly rolling out
actions in their policy parameterization, which can disagree with realized robot
motion due to controller scaling, frame conventions, orientation
parameterization, latency, tracking bandwidth, and load-dependent execution
error. Such mismatch can degrade both the target pose $x_t^g$ and the alignment
reward in Eq.~\eqref{eq:igrl_reward}. We therefore map each
imitation-predicted action chunk through three stages:
\begin{equation}
  \mathbf{a}_{t:t+T_a-1}
  \xrightarrow{\;g_{\mathrm{ctrl}}\;}
  \mathbf{u}_{t:t+T_a-1}
  \xrightarrow{\;f_\phi\;}
  \hat{\mathbf{u}}_{t:t+T_a-1}
  \xrightarrow{\;\Pi_{\mathrm{eef}}\;}
  \hat{x}_{t:t+T_a}.
\end{equation}

The first stage, $g_{\mathrm{ctrl}}$, applies known controller transformations,
including action rescaling, frame conversions, and the controller's orientation
parameterization. The second stage, $f_\phi$, is a lightweight learned residual
model fitted from demonstrations to predict the executed counterpart
$\hat{\mathbf{u}}$ of each controller command. This residual captures only
first-order, time-invariant tracking effects, such as latency, finite bandwidth,
and command attenuation; it is not intended to model contact dynamics, force
interactions, or task-specific object motion. The final stage,
$\Pi_{\mathrm{eef}}$, converts the predicted executed commands into an EEF
reference trajectory. In our experiments, where commands correspond to relative
EEF motions, $\Pi_{\mathrm{eef}}$ accumulates the predicted position deltas and
composes the orientation deltas from the current measured EEF pose. More
generally, $\Pi_{\mathrm{eef}}$ can be replaced by the identity map for absolute
EEF targets, or by forward kinematics for joint-space commands. We denote the
complete mapping as the composition
$F_{\mathrm{map}}=\Pi_{\mathrm{eef}}\circ f_\phi\circ g_{\mathrm{ctrl}}$.

At every chunk refresh, the trajectory is anchored to the current measured robot
state, so prediction errors remain local to the short guidance horizon rather
than accumulating over the full episode. This mapping should therefore be viewed
as a local geometric guide rather than a task dynamics model: it keeps early
exploration near imitation-guided behaviors, while online RL still optimizes the
true sparse task objective and can correct residual mapping errors through
interaction. We ablate the learned residual component by comparing TIGER with
and without $f_\phi$, isolating the effect of correcting systematic controller
mismatch beyond the analytic transformations in $g_{\mathrm{ctrl}}$. The mapping
is trained once per robot/controller setup and reused across tasks; details,
prediction-error analysis, and this ablation are provided in
\suppref{app:mapping}.

\subsection{Imitation-Guided Reward and Observation}
\label{subsec:imitation_guided_reward}
Inspired by prior work on imitation-based motion shaping~\cite{2018-TOG-deepMimic}, we design an imitation-guided reward and observation that encourage the RL policy to track a reference end-effector trajectory generated by an IL policy.

\begin{wrapfigure}{r}{0.55\textwidth}
  \vspace{-2\baselineskip}
  \begin{minipage}{\linewidth}
  \captionsetup{hypcap=false}
  \begin{algorithm}[H]
  \caption{TIGER imitation guidance}
  \label{alg:tiger_guidance}
  \small
  \begin{algorithmic}[1]
  \State Observe $s_t$, extract current EEF pose $x_t$, and query IL policy for chunk
  $\mathbf{a}_{t:t+T_a-1}^{\mathrm{IL}}$
  \State Map chunk to future reference poses
  $\hat{x}_{t+1:t+T_a}^{g}=F_{\mathrm{map}}(x_t,\mathbf{a}_{t:t+T_a-1}^{\mathrm{IL}})$
  \State Set target $x_t^g \leftarrow \hat{x}_{t+T_H}^{g}$
  \For{$h=0,\ldots,T_H-1$}
      \State Execute RL action $a_{t+h}\sim\pi_\theta(\cdot|s_{t+h},x_t^g)$
      \State Compute $r^{\mathrm{TIGER}}_{t+h}=m_{t+h+1}-m_{t+h}$
  \EndFor
  \State Repeat with a new IL prediction
  \end{algorithmic}
  \end{algorithm}
  \end{minipage}
  \vspace{-1.1\baselineskip}
\end{wrapfigure}

\noindent\textbf{Trajectory Generation and Reference Goals.}
Every $T_H \leq T_a$ steps, TIGER queries the IL policy for an action chunk of
length $T_a$ and maps it to an end-effector reference trajectory using the
controller-aware trajectory mapping. The target pose $x_t^g$ is selected as the
pose to be reached after the first $T_H$ predicted steps, i.e., the short-horizon
look-ahead goal. The RL policy then executes for $T_H$ environment steps with
this fixed target. After these $T_H$ steps, TIGER queries the IL policy again
and updates the target accordingly. Algorithm~\ref{alg:tiger_guidance}
summarizes the online guidance procedure.

\noindent\textbf{Pose Alignment Metric.}
We define a smooth pose-alignment measure between the current end-effector pose
$x_t = (p_t, q_t)$ and a target pose $x_t^g = (p_t^g, q_t^g)$:
\begin{equation}
\label{eq:pose_alignment}
m_t(x_t, x_t^g)
=
\exp\!\left(-\lambda_p \| p_t - p_t^g \|_2 \right)
\cdot
\exp\!\left(-\lambda_q \| q_t \ominus q_t^g \| \right).
\end{equation}
Here, $\| p_t - p_t^g \|_2$ denotes the Euclidean distance between positions, while
$q_t \ominus q_t^g \triangleq q_t \otimes (q_t^g)^{-1}$ denotes the
relative rotation. The product form is bounded in $(0,1]$, equals $1$ at
perfect alignment, and follows common exponential pose-alignment rewards
\cite{2018-TOG-deepMimic, danielle_exp_rew_expample}.

\noindent\textbf{Imitation-Guided Reward.}
Given a transition $(s'_t, a_t, s'_{t+1})$ with augmented state
$s'_t = (s_t, x_t^g)$, the imitation-guided reward is defined as
\begin{equation}
\label{eq:igrl_reward}
r'_t
=
\lambda_{\text{TIGER}} r^\text{TIGER}_t
+
\lambda_{\text{env}} \, r^\text{env}_t ,
\end{equation}
where
\begin{equation}
    \label{eq:rewar_igrl}
    r^\text{TIGER}_t = \bigl( m_{t+1} - m_t \bigr).
\end{equation}
The coefficients $\lambda_{\text{TIGER}}$ and $\lambda_{\text{env}}$ balance trajectory
tracking against task completion.
Using the difference term $(m_{t+1} - m_t)$ encourages directional progress along the
reference trajectory, discourages stalling near locally aligned poses, and mitigates
reward hacking.
Although RL exploration can introduce distributional shift relative to the imitation policy, action chunking provides short-horizon multi-step forecasts that help keep the reference
trajectory closer to the demonstration distribution. As training progresses, the RL policy tends to better align with the reference trajectories.

\noindent\textbf{Reward Scaling.}
We choose $\lambda_{\mathrm{env}} > \lambda_{\mathrm{TIGER}}$ so that dense
imitation guidance helps early exploration but successful task completion
dominates the return. In practice, we set the ratio so that the maximum
imitation-guided return from an unsuccessful rollout is smaller than the
terminal task reward:
\begin{equation}
\label{eq:igrl_return}
\sum_{t=0}^{T} \lambda_{\mathrm{TIGER}} r^{\mathrm{TIGER}}_t
\;\lesssim\;
\frac{1}{k}\lambda_{\mathrm{env}} r^{\mathrm{env}},
\qquad k \geq 1 .
\end{equation}
We use $k \approx 1.5$; sensitivity results are in
\suppref{app:ablation_k} and additional details are in \suppref{app:hp}.

\subsection{Offline Pretraining}
\label{subsec:pretraining_phase}

Before online fine-tuning, TIGER pretrains the actor and critic on
demonstrations to reduce early off-manifold exploration. For each
demonstration state, we compute an IL-guided target pose $x_t^g$ and form the
augmented state $s'_t=(s_t,x_t^g)$. For a candidate action $a_t$, the
controller-aware mapping predicts the next end-effector pose ($\hat{x}_{t+1}$)
which gives a predicted look-ahead TIGER signal without access to the true next observation
\begin{equation}
\label{eq:tilde_r_igrl}
    \tilde r^{\mathrm{TIGER}}_{t}(s'_t,a_t)
    =
    m_{t+1}(\hat{x}_{t+1}, x_t^g).
\end{equation}

We use this signal only inside the conservative regularizer, not in the TD
target. The pretraining objective is
\begin{equation}
\label{eq:pretraining_loss}
\begin{aligned}
\mathcal{L}_{\mathrm{pre}}(\theta)
&=
\mathbb{E}_{\mathcal{D}}
\Big[
\big(Q_\theta(s'_t,a_t)-y_t\big)^2
\Big] \\
&\quad
+\lambda_{\mathrm{cql}}
\mathbb{E}_{\mathcal{D}}
\Big[
\tau\log
\mathbb{E}_{a\sim\mu(\cdot\mid s'_t,s'_{t+1})}
\exp\!\big(w(s'_t,a)Q_\theta(s'_t,a)/\tau\big)
\Big] \\
&\quad
-\lambda_{\mathrm{cql}}
\mathbb{E}_{\mathcal{D}}
\big[Q_\theta(s'_t,a_t)\big].
\end{aligned}
\end{equation}
where $\mu(\cdot|s'_t,s'_{t+1})$ is a proposal distribution over actions and
\begin{equation}
\label{eq:td_target}
    y_t = r'_t + \gamma
    \mathbb{E}_{a\sim\pi_{\mathrm{RL}}(\cdot|s'_{t+1})}
    [Q_{\bar{\theta}}(s'_{t+1},a)]
\end{equation}
and
\begin{equation}
\label{eq:task_aware_weighting}
    w(s'_t,a)=1-\alpha\,\tilde r^{\mathrm{TIGER}}_{t}(s'_t,a),
    \qquad \alpha\in[0,1].
\end{equation}
The actor is pretrained to maximize the critic value
$Q_\theta(s'_t,\pi_\theta(s'_t))$ on augmented dataset states.

\noindent\textbf{Interpretation.}
The conservative term applies CQL to a task-modulated value
$\widetilde Q_\theta(s,a)=w(s,a)Q_\theta(s,a)$ while keeping the Bellman target
unchanged. Thus, TD learning estimates values from real transitions, while the
conservative penalty suppresses OOD actions during pretraining. Unlike standard
CQL, which penalizes sampled OOD actions, TIGER relaxes the penalty for
actions locally consistent with the IL reference and remains conservative
elsewhere. This biases pretraining toward task-aligned directions without adding
model-predicted rewards to the TD backup. It reduces to standard CQL when
$\alpha=0$; additional details are given in \suppref{app:goal_aware_cql}.

\section{Experiments}
\label{sec:Experiments}
We evaluate TIGER on simulated and real-robot sparse-reward manipulation tasks.
Our experiments ask whether TIGER
(i) accelerates sparse-reward learning and improves beyond its BC prior,
(ii) outperforms learned reward alternatives under a matched RL backend,
(iii) retains its benefit in a multi-task benchmark,
and (iv) improves real-robot learning efficiency and measured safety.

We instantiate TIGER with two off-policy RL backends, RLPD~\cite{rlpd}
and IBRL~\cite{ibrl}. We compare against representative sparse-reward robot
learning methods that use demonstrations, pretrained imitation policies, or
dense surrogate rewards: RLPD, which mixes prior demonstrations with online
replay; IBRL, which uses imitation for action proposal and value bootstrapping;
LaNE~\cite{LaNE}, which constructs dense rewards by aligning observations with
expert trajectories in visual representation space; DSRL~\cite{dsrl}, which
improves a frozen flow-matching policy through latent-space RL; and
ResFiT~\cite{resfit}, which fine-tunes behavior cloning through residual
action-space corrections. We use flow-matching~\cite{flow-matching} behavior cloning as our BC policy.
For the targeted learned-reward comparison, we additionally evaluate
ReWiND~\cite{rewind}, which learns language-conditioned progress rewards from demonstrations,
and Robometer~\cite{liang2026robometer}, a general-purpose video-language reward
model trained with frame-level progress and trajectory-comparison supervision.
Full baseline descriptions, architecture choices,
and hyperparameters are provided in \suppref{app:baseline_details}, \suppref{app:hp} and \suppref{app:network_architectures}.

\subsection{Simulation}
\label{subsec:simulation}
\noindent\textbf{Setup.} We evaluate TIGER on three MetaWorld tasks~\cite{metaworld}
(Assembly, Box Close, and Stick Pull), three tasks from the Robomimic
benchmark~\cite{robomimic2021} (Can, Square, and ToolHang),
and a humanoid shape-sorting insertion task. The tasks span sparse-reward manipulation problems with varying horizons and
contact precision. ToolHang requires up to 600 control
steps and precise insertion, while the humanoid task further introduces tight
insertion tolerances. We use three demonstrations
per MetaWorld task from~\cite{ibrl}, 20 demonstrations for Can, 150 for
ToolHang, and 100 for each of Square and the humanoid task for BC
training. For the learned-reward comparison, we train ReWiND and fine-tune
Robometer on the same demonstrations as TIGER and use RLPD as the common
backend on Assembly, Box Close, Stick Pull, and Square.

We additionally evaluate LIBERO-Spatial~\cite{liu2023libero} with 10 and 50
demonstrations per task. In each regime, we train one shared multi-task BC
policy and one shared multi-task RL policy per online method across all ten
tasks, using one-hot task conditioning. Training and evaluation use the same
ten tasks. 
All experiments are run over five seeds. Full details are provided in
Appendix~\ref{app:simulation_details}.
\begin{table*}[t]
  \centering
  \caption{
  Simulation benchmark summary. Each cell reports $S_{50}\downarrow/S_{90}\downarrow/F\uparrow$,
  where $S_{50}$ and $S_{90}$ are environment steps, in thousands, required to
  reach 50\% and 90\% success, and $F$ is final success. BC has no online-learning thresholds otherwise ``--'' denotes that the
  threshold was not reached.
  }
  \label{tab:sim_summary}
  \resizebox{\textwidth}{!}{
  \begin{tabular}{lccccccc}
  \toprule
  &
  \multicolumn{3}{c}{Robomimic} &
  \multicolumn{3}{c}{MetaWorld} &
  \multicolumn{1}{c}{Humanoid} \\
  \cmidrule(lr){2-4}
  \cmidrule(lr){5-7}
  \cmidrule(lr){8-8}
  Method
  & Can & Square & ToolHang
  & Assembly & Box Close & Stick Pull
  & Shape Sorting \\
  \midrule
  BC
  & --/--/0.7
  & --/--/0.6
  & --/--/0.5
  & --/--/0.7
  & --/--/0.38
  & --/--/0.25
  & --/--/0.1 \\
  
  LaNE
  & --/--/0.25
  & --/--/0.025
  & --/--/0.0
  & --/--/0.08
  & --/--/0.06
  & 45/--/0.6
  & --/--/0.0 \\
  
  RLPD
  & 55/--/0.71
  & 120/--/0.79
  & --/--/0.0
  & 40/50/0.95
  & 35/45/0.94
  & --/--/0.18
  & --/--/0.0 \\
  
  IBRL
  & 35/75/0.91
  & 80/160/0.91
  & 470/--/0.0
  & 25/30/0.96
  & 25/40/0.94
  & 35/--/0.72
  & 115/--/0.78 \\
  
  DSRL
  & \textbf{5}/--/0.64
  & --/--/0.37
  & --/--/0.0
  & \textbf{5}/--/0.73
  & --/--/0.3
  & --/--/0.12
  & --/--/0.0 \\
  
  ResFiT
  & 15/--/0.81
  & 90/--/0.7
  & \textbf{200}/--/0.25
  & 10/65/0.88
  & \textbf{10}/--/0.55
  & --/--/0.30
  & --/--/0.38 \\
  
  TIGER + RLPD
  & 25/50/0.95
  & 90/160/0.87
  & 430/--/0.21
  & 25/30/0.95
  & 25/45/0.93
  & 40/50/\textbf{0.92}
  & 135/--/0.77 \\
  
  TIGER + IBRL
  & 20/\textbf{50}/\textbf{0.96}
  & \textbf{50}/\textbf{150}/\textbf{0.92}
  & 330/\textbf{600}/\textbf{0.79}
  & 15/\textbf{20}/\textbf{0.96}
  & 20/\textbf{40}/\textbf{0.94}
  & \textbf{25}/\textbf{45}/0.9
  & \textbf{50}/\textbf{205}/\textbf{0.93} \\
  \bottomrule
  \end{tabular}
  }
  \vspace{-0.5cm}
  \end{table*}

\noindent\textbf{Results and Discussion}
Table~\ref{tab:sim_summary} summarizes simulation performance using
sample-efficiency thresholds and final success. To isolate online
imitation-guided rewards, we disable pretraining for all MetaWorld experiments.
Across tasks, TIGER primarily improves early learning over the matched RL
backend: TIGER+IBRL reaches $S_{50}$ sooner than IBRL on every task and reaches
$S_{90}$ on ToolHang and shape sorting, where IBRL does not. Final success is
generally matched or improved, with the largest gains on tasks where sparse
exploration is the bottleneck, such as ToolHang, Stick Pull, and shape sorting.
This pattern supports TIGER's central mechanism: task-space guidance is most
useful when imitation provides a reasonable short-horizon direction, but pure RL
struggles to discover successful contact-rich behavior.
\begin{wrapfigure}{r}{0.45\textwidth}
\vspace{-\baselineskip}
  \centering
  \includegraphics[width=\linewidth]{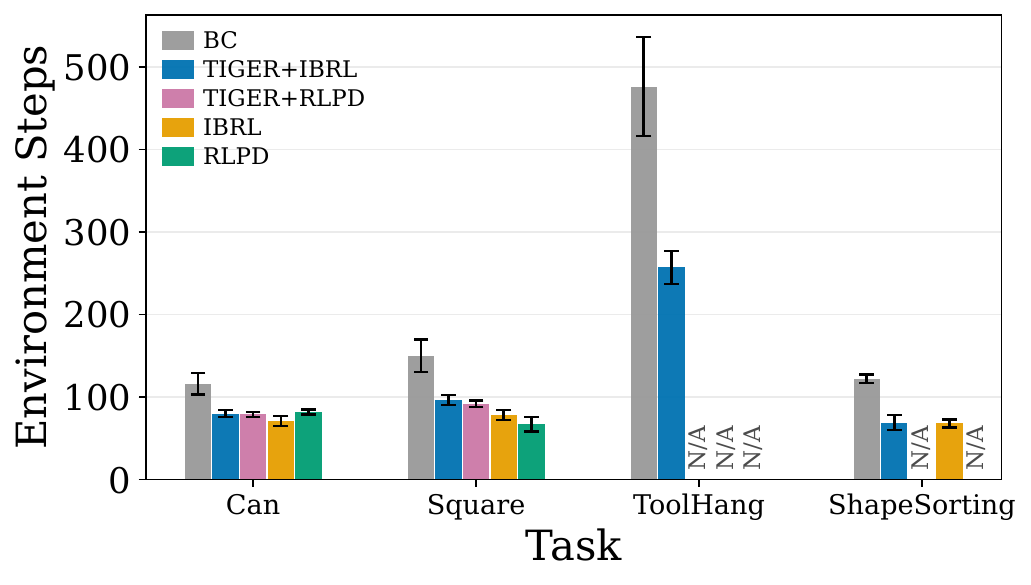}
  \vspace{-0.6cm}
  \caption{Final episode length achieved by each method.}
  \label{fig:episode_length}
  \vspace{-0.55cm}
\end{wrapfigure}
The hardest tasks also test improvement beyond weak or imperfect BC priors. On
Shape Sorting, BC achieves only $10\%$ final success, whereas TIGER+IBRL reaches
$S_{90}$ at $205\,\mathrm{k}$ steps and $93\%$ final success; most BC failures
occur at the final insertion, so earlier action chunks can still provide useful
local guidance. On ToolHang, BC reaches $50\%$ final success, compared with
$79\%$ for TIGER+IBRL. Fig.~\ref{fig:episode_length} further reports shorter final episodes than BC
on the challenging tasks in these benchmarks, and \suppref{app:extended_ablation} shows strong
performance with more variable multi-human demonstrations. Because online
fine-tuning imposes no explicit action-support or KL constraint tying the RL
policy to BC, these results demonstrate empirical improvement beyond---though
not formal independence from---the imitation prior.
Full training curves are provided in
\suppref{app:sim_training_curves}.

\noindent\textbf{Comparison with learned reward models.}
To compare dense-guidance mechanisms while controlling demonstrations and the RL
backend, we evaluate TIGER, ReWiND, and Robometer with the same demonstrations
and RLPD backend over five seeds on three MetaWorld tasks and Robomimic Square
(Table~\ref{tab:additional-simulation}, panel~(a)). TIGER is strongest overall,
with clear gains on StickPull and Square, supporting task-space guidance over these
learned visual rewards on the evaluated tasks.

\noindent\textbf{Multi-task LIBERO-Spatial.}
We evaluate two demonstration regimes: 10 and 50 demonstrations per task (100
and 500 demonstrations in total, respectively). For each regime, we train one
shared multi-task BC policy and one shared multi-task RL policy per online
method across all ten LIBERO-Spatial tasks, using one-hot task conditioning and
five seeds. Across both demonstration regimes, TIGER+IBRL reaches $S_{50}$ and $S_{90}$
earlier than IBRL and improves final success by two percentage points. BC
achieves final success of $0.80$ and $0.91$ with 10 and 50 demonstrations per
task, respectively (Table~\ref{tab:additional-simulation}, panel~(b)).
\begin{table}[t]
  \centering
  \caption{Additional simulation comparisons. Entries report
  $S_{50}\!\downarrow/S_{90}\!\downarrow/F\!\uparrow$; environment steps are in
  thousands and results are averaged over five seeds. Panel (a) uses the same
  demonstrations and RLPD backend. Panel (b) reports task averages over all ten
  LIBERO-Spatial tasks; BC has no online-learning thresholds. A dash otherwise
  indicates that a threshold was not reached.}
  \label{tab:additional-simulation}
  \medskip
  \begin{minipage}[t]{0.55\linewidth}
    \centering
    \textbf{(a) Learned rewards (RLPD)}

    \smallskip
    {\setlength{\tabcolsep}{2pt}
    \begin{tabular}{@{}lccc@{}}
      \toprule
      Task & ReWiND & Robometer & TIGER \\
      \midrule
      Assembly   & \textbf{25/30/}.95  & \textbf{25/30/.95} & \textbf{25/30/.95} \\
      Box Close  & 30/60/.87  & 25/50/.92          & \textbf{25/45/.93} \\
      Stick Pull & 50/60/.84  & 55/--/.60          & \textbf{40/50/.92} \\
      Square     & 130/--/.74 & 120/--/.71         & \textbf{90/160/.87} \\
      \bottomrule
    \end{tabular}}
  \end{minipage}\hfill
  \begin{minipage}[t]{0.43\linewidth}
    \centering
    \textbf{(b) Multi-task LIBERO-Spatial}

    \smallskip
    {\setlength{\tabcolsep}{2pt}
    \begin{tabular}{@{}lccc@{}}
      \toprule
      Demos/task & BC & IBRL & TIGER \\
      \midrule
      10 & --/--/.80 & 80/140/.93 & \textbf{60/100/.95} \\
      50 & --/--/.91 & 80/140/.94 & \textbf{50/80/.96} \\
      \bottomrule
    \end{tabular}}
  \end{minipage}
\end{table}

\subsection{Real Robot Experiments}
\label{subsec:real_robot_experiments}
\begin{figure*}[t!]
  \centering

  \makebox[\textwidth][c]{%
    \begin{minipage}{0.3\textwidth}
      \centering
      \text{Block Picking}
    \end{minipage}
    \hfill
    \begin{minipage}{0.3\textwidth}
      \centering
      \text{Drawer Opening}
    \end{minipage}
    \hfill
    \begin{minipage}{0.3\textwidth}
      \centering
      \text{Rotary Insertion}
    \end{minipage}
  }

  \begin{subfigure}[b]{0.3\textwidth}
    \centering
    \includegraphics[width=\textwidth]{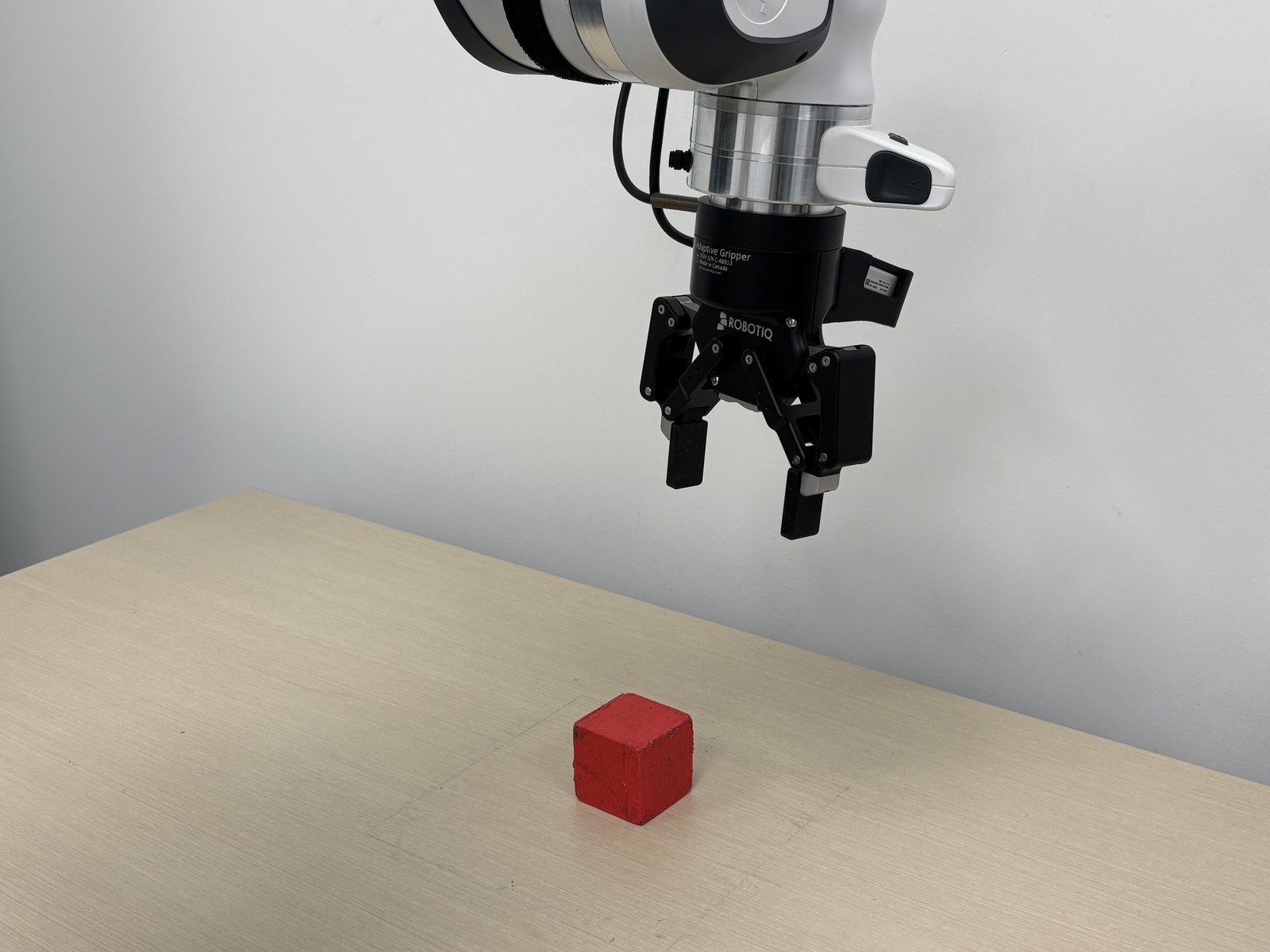}
  \end{subfigure}
  \hfill
  \begin{subfigure}[b]{0.3\textwidth}
    \centering
    \includegraphics[width=\textwidth]{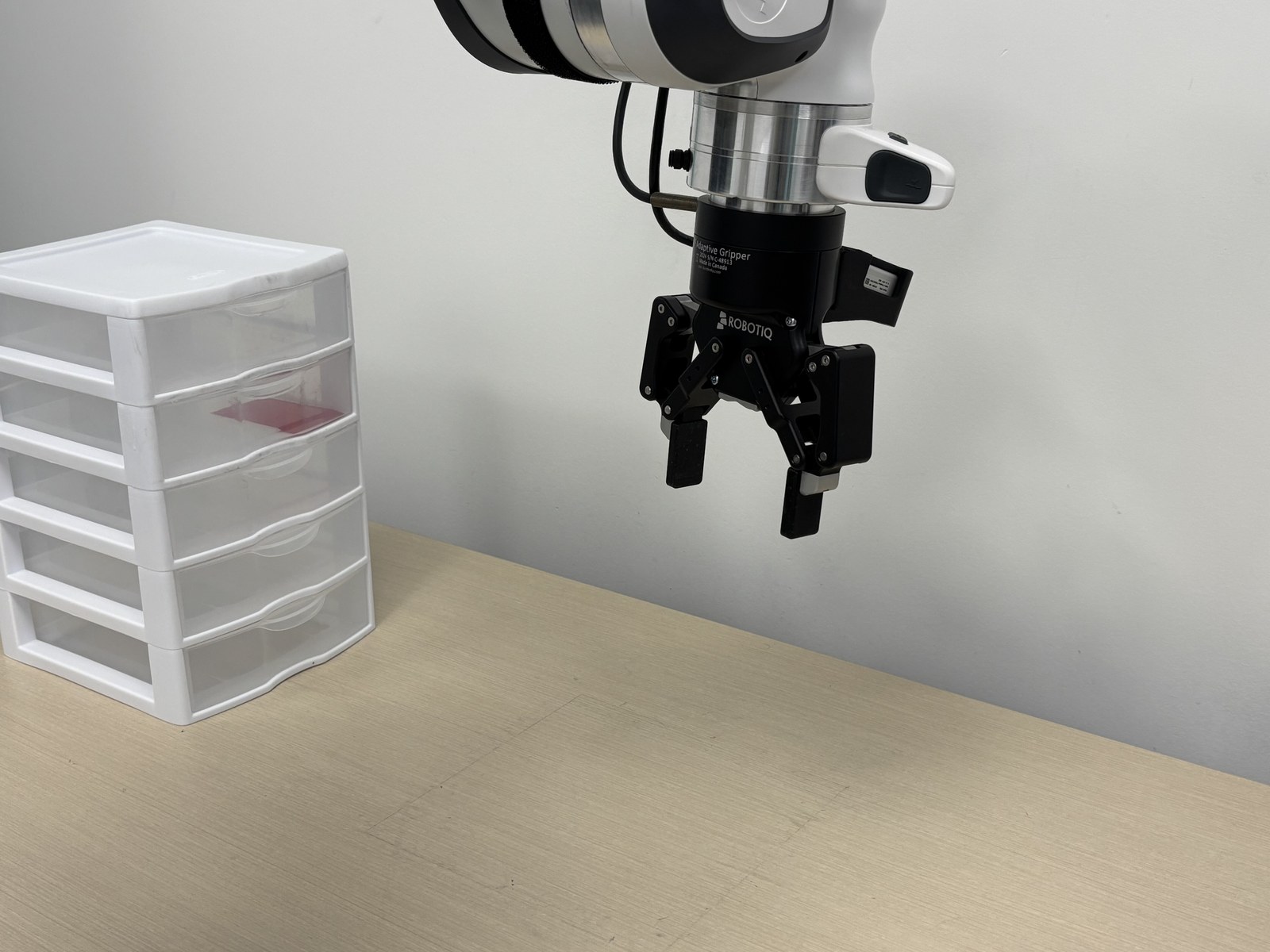}
  \end{subfigure}
  \hfill
  \begin{subfigure}[b]{0.3\textwidth}
    \centering
    \includegraphics[width=\textwidth]{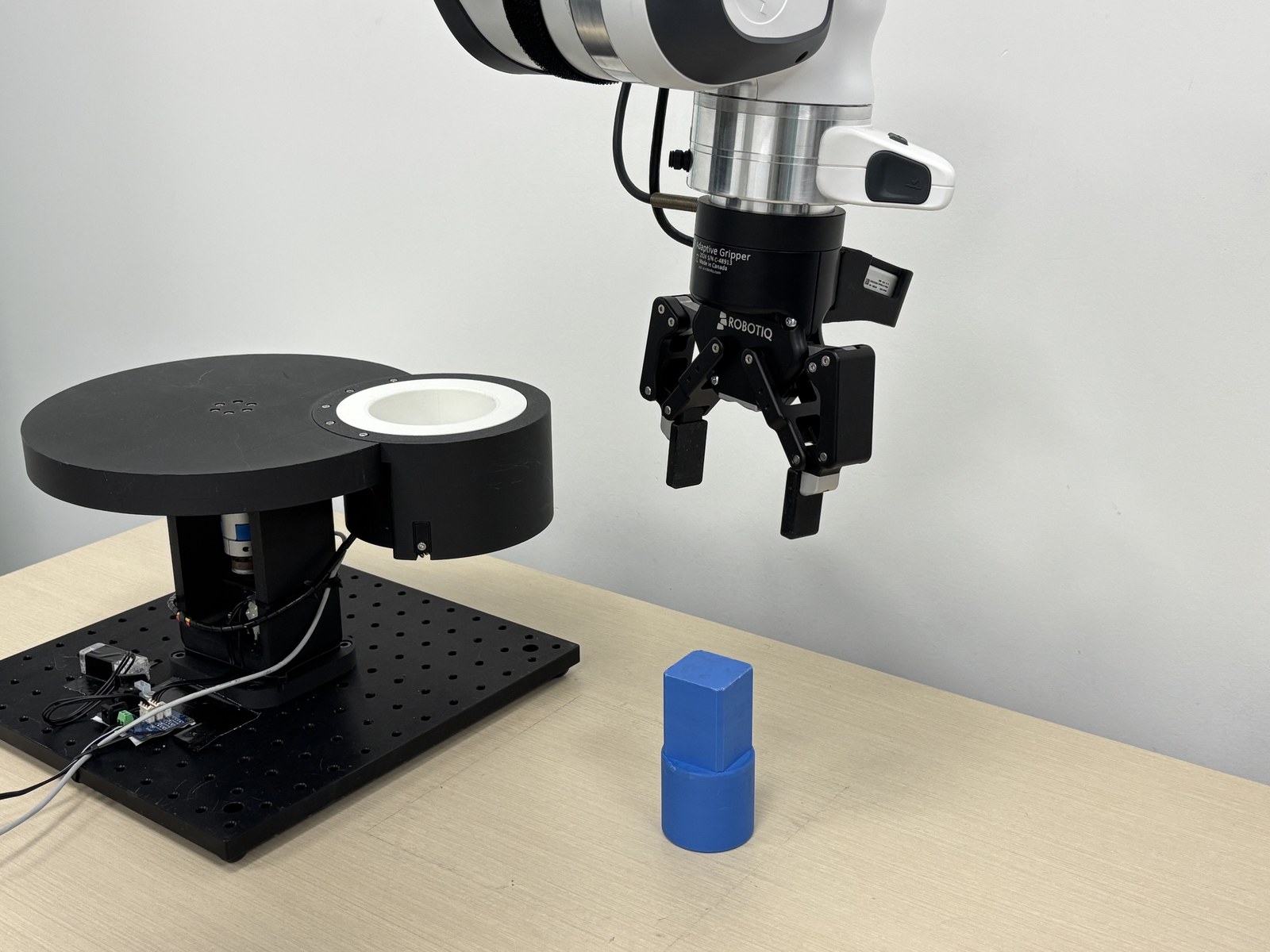}
  \end{subfigure}

  \begin{subfigure}[b]{0.33\textwidth}
    \centering
    \includegraphics[width=\textwidth]{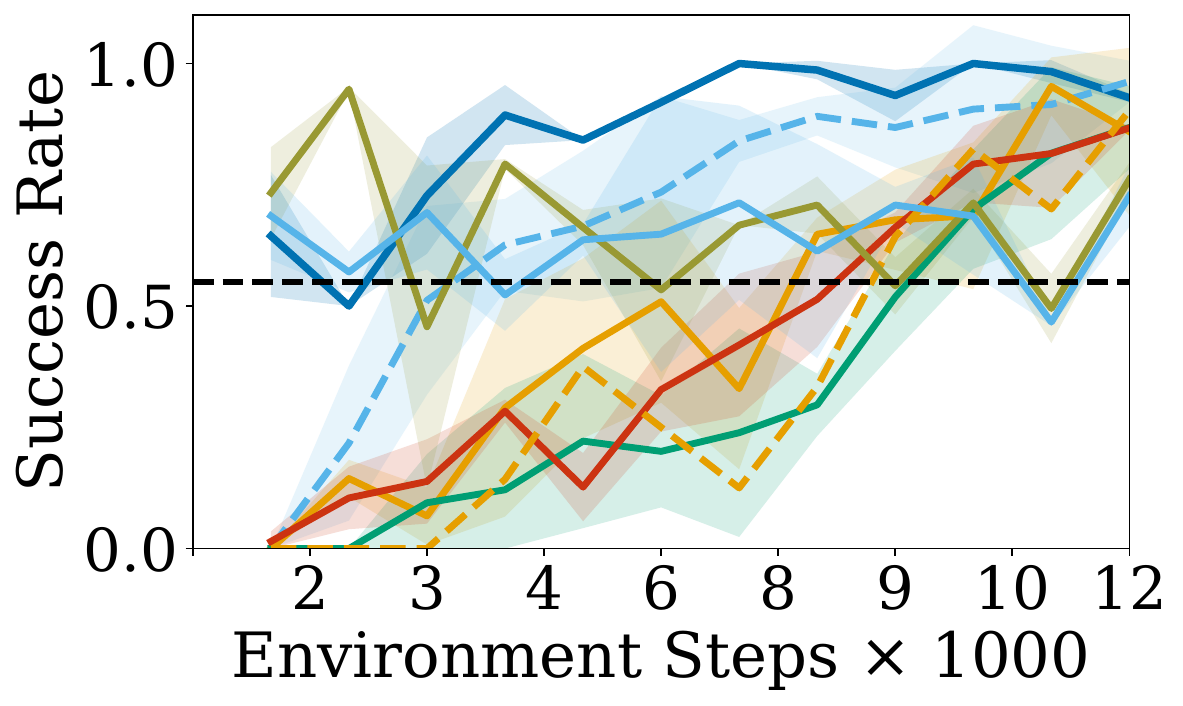}
  \end{subfigure}
  \hfill
  \begin{subfigure}[b]{0.32\textwidth}
    \centering
    \includegraphics[width=\textwidth]{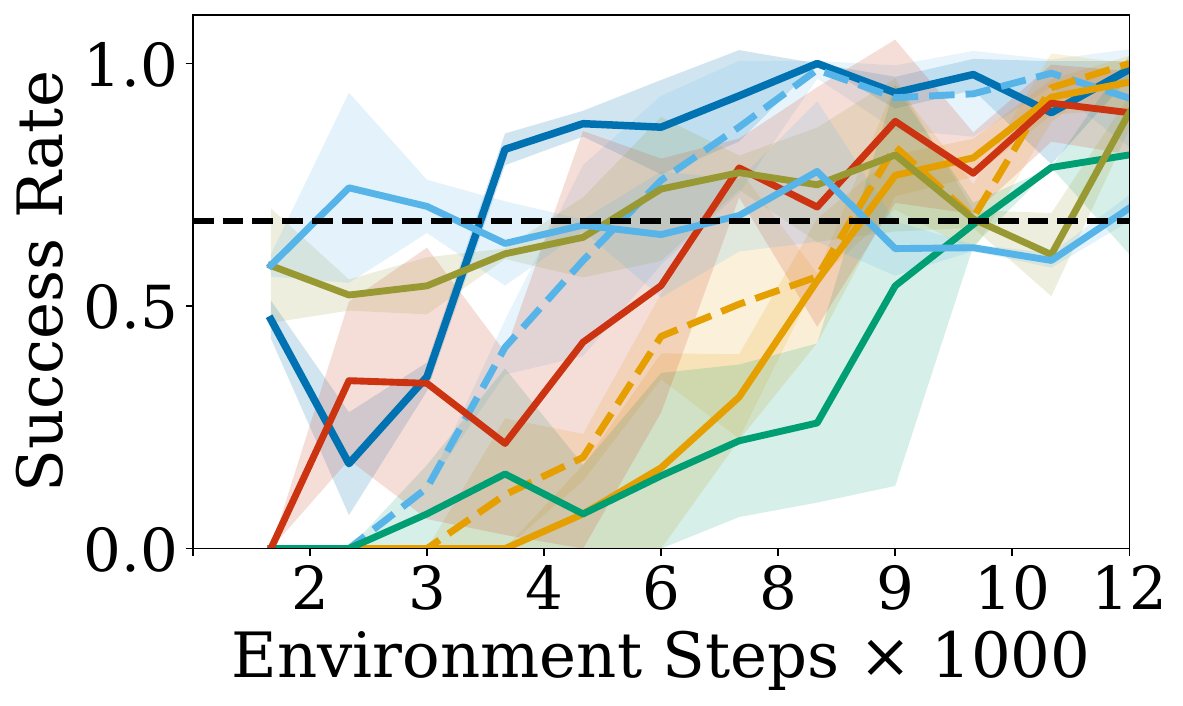}
  \end{subfigure}
  \hfill
  \begin{subfigure}[b]{0.33\textwidth}
    \centering
    \includegraphics[width=\textwidth]{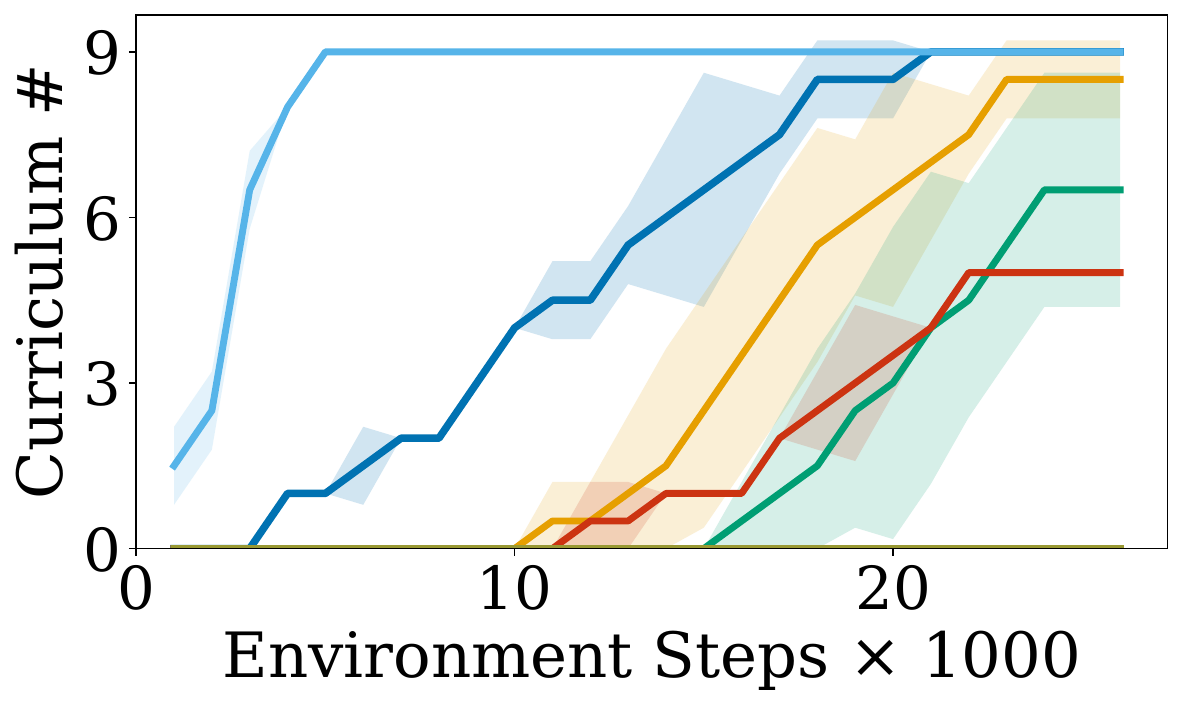}
  \end{subfigure}
    \vspace{0.4cm}
  \begin{subfigure}[b]{0.99\textwidth}
    \centering
    \includegraphics[width=\textwidth]{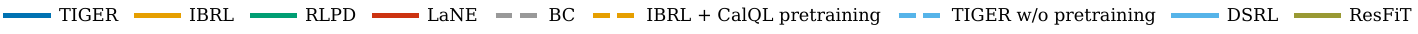}
  \end{subfigure}
  \vspace{-0.5cm}
\caption{\textbf{Top:} Visualization of each task. \textbf{Bottom:} Training curves for each task. The x-axis represents environment steps in intervals of 1000. For block picking and drawer opening, the y-axis indicates the average task success rate over each 1000-step interval. For rotary insertion, the y-axis denotes the curriculum stage. Task completion requires reaching stage \#9.}
\label{fig:real_franka_exp}
\vspace{-0.3cm}
\end{figure*}
\begin{figure}[t]
  \centering
  \includegraphics[width=0.45\linewidth]{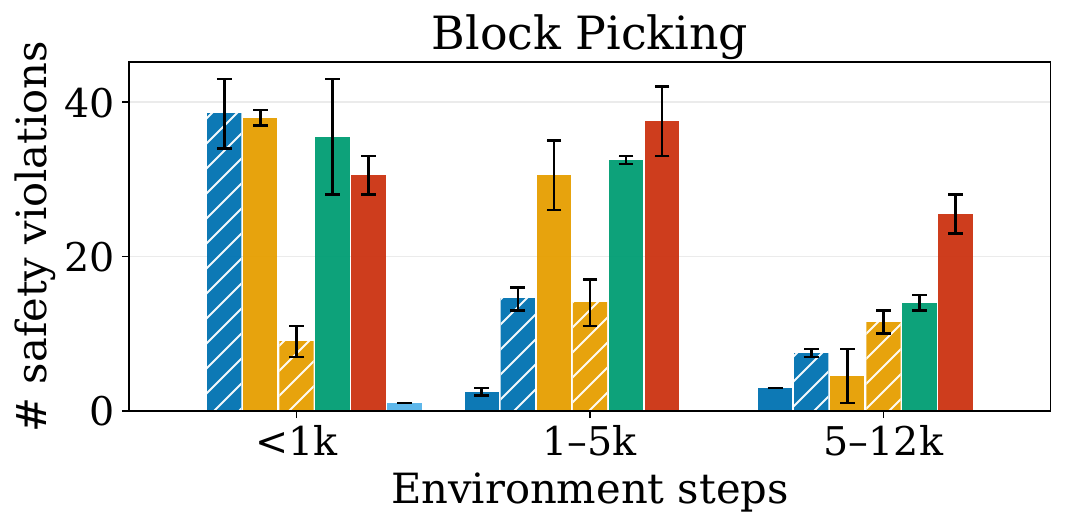}\hfill
  \includegraphics[width=0.45\linewidth]{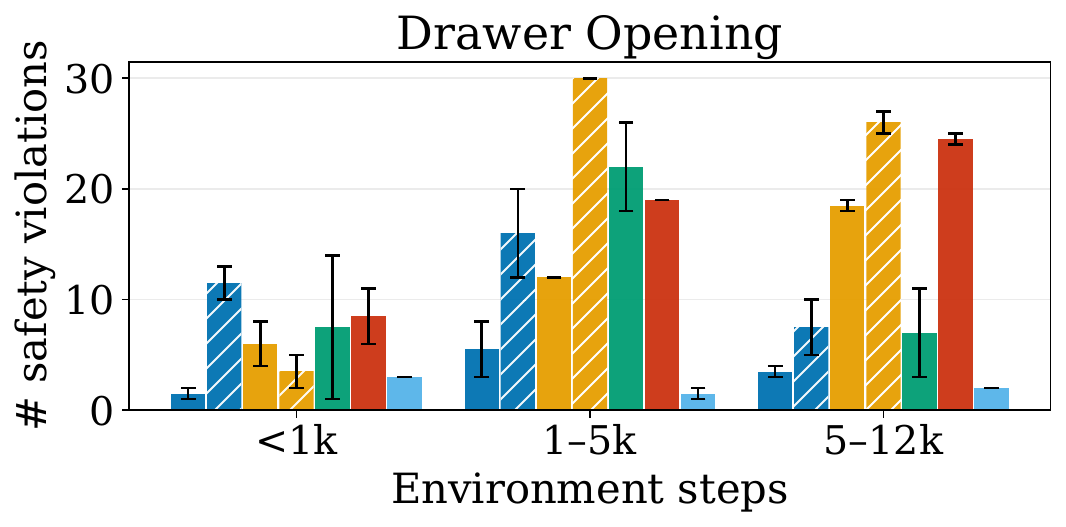}\\
  \includegraphics[width=0.8\linewidth]{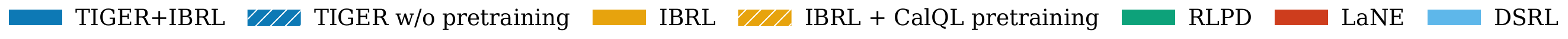}
  \caption{Safety violations on the two real-robot tasks (block picking and drawer opening). Bars show the mean number of episode resets across two seeds with min--max ranges.}
  \label{fig:safety_violation_analysis}
  \vspace{-0.4cm}
\end{figure}
\noindent\textbf{Task Setup.}
We evaluate TIGER on three contact-rich real-robot manipulation tasks:
\emph{Block Picking}, \emph{Drawer Opening}, and \emph{Rotary Insertion}, using
a Franka Emika Research 3 robot (Fig.~\ref{fig:real_franka_exp}). Based on the
simulation results, we instantiate TIGER with IBRL~\cite{ibrl} as the RL backend.
We use 10 teleoperated demonstrations for \emph{Block Picking}, 10 for \emph{Drawer Opening}, and 50
for \emph{Rotary Insertion}. All methods are trained on the real robot for the same number of environment steps, with two independent training runs (seeds) per method. We additionally train TIGER from scratch on \emph{Block Picking}
and \emph{Drawer Opening} to ablate pretraining.

The tasks cover grasping, articulated-object interaction, and dynamic insertion.
\emph{Block Picking} requires grasping a randomly placed block;
\emph{Drawer Opening} requires opening the correct small-handled drawer, with
zero reward for opening the wrong drawer; and \emph{Rotary Insertion} requires 
inserting an object into a randomized rotating slot
through a staged curriculum shared by all methods under the same progression
rule, ending in full-circle randomization and continuous rotation. Full hardware,
controller, randomization, reward, and curriculum details are provided in
\suppref{app:real_detail}.

\noindent\textbf{Results and Discussion.}
The bottom row of Fig.~\ref{fig:real_franka_exp} shows real-robot training
curves. For \emph{Block Picking} and \emph{Drawer Opening}, we report average
episode success every 1k environment steps; for \emph{Rotary Insertion}, we
report curriculum stage, which better captures progress on this long-horizon
task. On \emph{Block Picking} and \emph{Drawer Opening}, TIGER converges in
approximately 8k and 7k steps, respectively, versus roughly 12k steps for the
baselines. TIGER also starts with nonzero success due to imitation-guided
pretraining and incurs fewer early safety violations, although we observe a
temporary drop during offline-to-online transition, consistent with prior
offline-to-online RL~\cite{corl_offline_rl}.

\begin{wraptable}{r}{0.5\textwidth}
  \vspace{-10pt}
  \caption{Rotary insertion evaluation.}
  \centering
  \small
  \begin{tabular}{l|c|c}
  \hline
                & Stationary & Rotating \\[-3pt]
                & Random Location & Location \\ \hline
  BC           & 60\%        & 10\%      \\
  TIGER        & 100\%       & 90\%      \\
  IBRL         & 90\%        & 77.5\%    \\
  DSRL         & 60\%        & 20\%      \\ \hline
  \end{tabular}
  \vspace{-10pt}
  \label{table:rotary_insertion_evaluation}
\end{wraptable}
\emph{Rotary Insertion} is more challenging: the agent must pick up the object,
track a moving target, and insert it while the socket rotates. All online methods
use the same staged curriculum and progression rule. The demonstrations contain
only static, randomly oriented disks; the demonstrator never tracks a moving
target. Despite this mismatched prior, TIGER reaches the final curriculum stage
in both seeds within the fixed budget of $26\,\mathrm{k}$ environment steps
($\approx 3$ hours per run), whereas IBRL reaches it in only one of two seeds;
RLPD, LaNE, and ResFiT make limited progress. DSRL's rapid progress through the
early static stages largely reflects behavior inherited from its frozen BC
prior.

Table~\ref{table:rotary_insertion_evaluation} reports final performance in the
last two curriculum stages for BC, TIGER, DSRL, and the best-performing IBRL
seed. BC achieves $60\%$ success in the stationary-random setting but only
$10\%$ under rotation. DSRL matches BC at $60\%$ in the stationary-random
setting and reaches $20\%$ under rotation, compared with $90\%$ for TIGER.
Thus, TIGER achieves the highest success in the most challenging rotating
setting.
\begin{figure}[t]
  \centering
  \captionsetup{hypcap=false}
  \begin{minipage}[t]{0.48\linewidth}
      \centering
      \includegraphics[width=\linewidth]{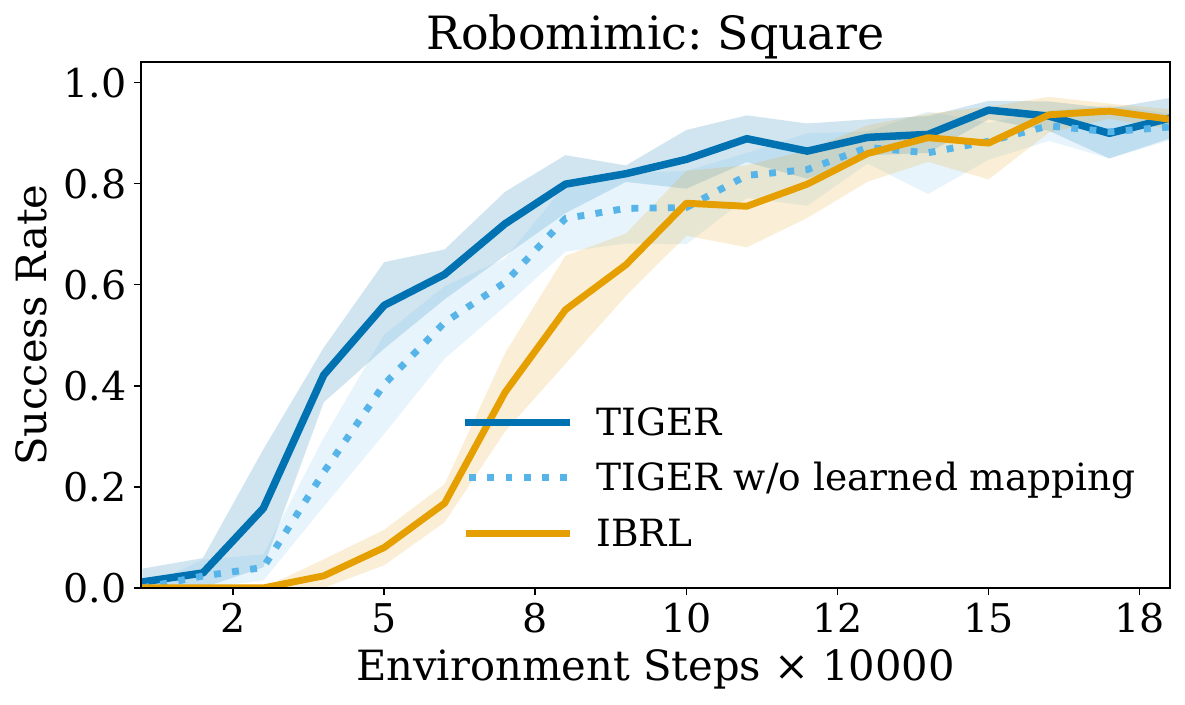}
      \vspace{-0.5cm}
      \caption{Learned controller-residual ablation.}
      \label{fig:ablation_controller_robosuite}
  \end{minipage}\hfill
  \begin{minipage}[t]{0.48\linewidth}
      \centering
      \includegraphics[width=\linewidth]{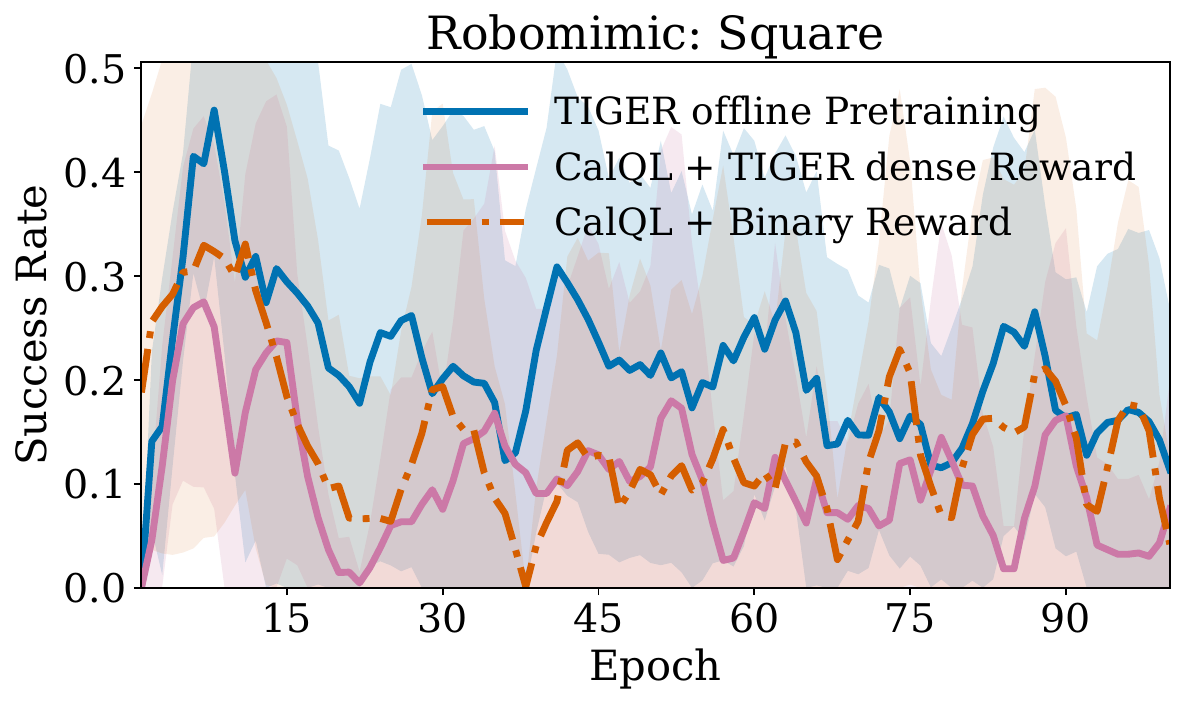}
      \vspace{-0.5cm}
      \caption{Pretraining objective ablation.}
      \label{fig:offline_robosuite}
  \end{minipage}
  \vspace{-0.7cm}
\end{figure}
\subsection{Ablation Study}
\label{subsec:ablation}
We analyze the learned mapping residual, offline pretraining, and measured
safety violations to assess their contributions to TIGER's performance. Additional ablations on demonstration quality, guidance
design, and mapping transfer are in \suppref{app:extended_ablation}.

\noindent\textbf{Effect of pretraining.}
To isolate pretraining, we repeat \emph{Block Picking} and \emph{Drawer Opening}
with TIGER from scratch (Fig.~\ref{fig:real_franka_exp}).
TIGER without pretraining still outperforms IBRL, showing that the imitation-guided reward is
beneficial on its own. However, pretraining improves early exploration and
reduces boundary violations. We also pretrain IBRL with CalQL and observe no meaningful
gain in performance or boundary violations, suggesting that generic offline RL
pretraining is less aligned with the task objective.

\noindent\textbf{Safety analysis.}
We log unsafe events during real-robot training on \emph{Block Picking} and
\emph{Drawer Opening} (Fig.~\ref{fig:safety_violation_analysis}).
A violation is leaving the task-specific safety box or
triggering a controller failure; \suppref{app:real_detail} gives the full bounds and reset
conditions. TIGER pretraining substantially reduces early resets by keeping
behavior closer to the demonstration manifold. Removing pretraining increases
exploration and resets, whereas sparse-reward IBRL incurs fewer early violations
partly because it engages the task more slowly. DSRL likewise incurs few
violations under identical safety boxes and reset rules, consistent with
conservative behavior inherited from its frozen BC prior; that conservatism can
limit improvement beyond the prior. CalQL-pretrained IBRL does not yield similar
safety gains, indicating that TIGER's effect is not due to value initialization
alone.

\noindent\textbf{Effect of the learned controller residual.}
We ablate only the learned residual in the controller-aware trajectory mapping
on Square, leaving the known controller transformations $g_{\mathrm{ctrl}}$ and
EEF trajectory accumulation $\Pi_{\mathrm{eef}}$ active. The residual captures
systematic execution mismatch, including latency, action smoothing, and finite
tracking bandwidth, rather than contact dynamics. As shown in
Fig.~\ref{fig:ablation_controller_robosuite}, it
improves training speed, but TIGER without the residual still learns faster than
IBRL. In this Square ablation, short-horizon task-space guidance provides the
main gain, while the learned residual improves execution alignment and training

\noindent\textbf{Choice of pretraining objective.}
We compare three offline pretraining variants on the Square task:
CalQL with sparse rewards, CalQL with the TIGER dense reward, and full TIGER
pretraining. All variants use the same dataset from Sec.~\ref{sec:Experiments};
policies are evaluated over 100 rollouts per epoch, and
Fig.~\ref{fig:offline_robosuite} reports a moving average with window 4. TIGER
pretraining yields higher offline success and more stable learning. Standard
CalQL often degrades during training, consistent with prior observations in
offline robotic RL~\cite{kumar2021workflow}; adding the dense reward partially
mitigates this degradation, while the full TIGER objective gives the most
robust gains.
\section{Limitations}
TIGER assumes that short-horizon end-effector motion is an informative
local proxy for task progress, which is appropriate for the reaching,
grasping, insertion, and articulated-object manipulation tasks studied here.
This assumption may be weaker when success depends primarily on force
regulation, tactile feedback, deformable-object dynamics, dexterous in-hand
manipulation, hidden object state, or contact-mode reasoning rather than
end-effector pose. TIGER also relies on locally meaningful imitation
guidance: for multimodal demonstrations, the imitation policy selects one
short-horizon reference at each replanning step, providing locally consistent
guidance rather than enforcing global trajectory matching; however,
systematically misleading IL predictions can still bias the shaped reward
and target-conditioned observation. Finally, TIGER inherits the standard offline-to-online challenge that a pretrained policy may transiently degrade when exposed to online exploration; improving this handoff is an important direction for future work.
\clearpage
\acknowledgments{This work was conducted at and supported by Sanctuary AI, using its robotic and computational infrastructure.}

\bibliography{main}  %
\clearpage

\ifappendix
\section*{Appendix}
\appendix
\section{Experiment Details}
\label{app:experiment_details}
\subsection{Baseline Details}
\label{app:baseline_details}
We choose baselines that are most relevant to TIGER's setting:
sparse-reward robot learning with access to a small number of expert
demonstrations or a pretrained imitation policy. Specifically, we compare
against methods that incorporate demonstrations into RL, use demonstrations
to construct auxiliary guidance signals, or improve a frozen behavior-cloning
policy through online RL. In both simulation and real-robot experiments, we use
a flow-matching BC policy as the imitation backbone and report its standalone
performance as an imitation-only baseline. This comparison set allows us to
isolate whether TIGER's gains come from controller-aware task-space guidance,
rather than simply from access to demonstrations, imitation-policy
initialization, or residual policy improvement.

\noindent\textbf{IBRL:}
Imitation Bootstrapped Reinforcement Learning (IBRL)~\citep{ibrl} is our
closest baseline. IBRL first trains an imitation policy from demonstrations and
then uses it during online RL to propose candidate actions for exploration and
to bootstrap target values in critic updates. We include IBRL because it tests
the standard use of imitation as an action prior. In contrast, TIGER does not
treat the imitation policy as an action proposal mechanism; it converts
imitation-predicted action chunks into controller-aware task-space references
and rewards RL for progress toward those references.

\noindent\textbf{RLPD:}
Reinforcement Learning with Prior Data (RLPD)~\citep{rlpd} is a strong
sample-efficient off-policy RL baseline for online learning with offline data.
RLPD incorporates demonstrations through symmetric replay sampling, drawing
training batches from both the online replay buffer and the offline dataset,
and uses critic stabilization such as Layer Normalization to mitigate value
over-extrapolation. It therefore tests whether retaining demonstrations as
replay data is sufficient for sparse-reward online RL, without explicit
imitation objectives, auxiliary rewards, or task-space reference construction.
Whenever possible, RLPD uses the same observation space, action space, and
off-policy learner as TIGER, so performance differences primarily reflect the
effect of TIGER's imitation-guided task-space reward.

\noindent\textbf{LaNE:}
Latent Nearest-demonstration-guided Exploration (LaNE)~\citep{LaNE} constructs dense rewards by aligning observations with expert demonstrations in a pretrained visual representation space. We include LaNE because it is a recent sparse-reward visual RL method that also uses demonstrations to provide dense guidance. The comparison is important because LaNE represents the alternative design choice of reward shaping in image/latent observation space, whereas TIGER shapes reward in end-effector task space. This directly tests whether task-space imitation guidance is more suitable for contact-rich manipulation than visual trajectory alignment.

\noindent\textbf{DSRL:}
Diffusion Steering via Reinforcement Learning (DSRL)~\citep{dsrl} is a recent method for improving pretrained diffusion behavior cloning policies through online RL. Rather than fine-tuning the diffusion policy weights directly, DSRL treats the diffusion policy as a fixed black-box prior and learns to steer its latent noise space toward higher-reward actions. We include DSRL because it is closely aligned with our setting: it improves a frozen imitation policy using RL while preserving the
pretrained BC prior. The comparison highlights a key difference between the two approaches. DSRL uses RL to steer the action distribution of a frozen diffusion policy, whereas TIGER uses the imitation policy as a task-space guidance mechanism for a separate RL agent through controller-aware reference trajectories, dense progress rewards, and offline pretraining.

\noindent\textbf{ResFiT:}
ResFiT~\citep{resfit} is a residual off-policy RL fine-tuning method for
behavior-cloning policies. It keeps a pretrained BC policy as the base
controller and trains an RL residual to correct its actions. We include ResFiT
because it is a strong and conceptually relevant baseline for improving
imitation policies with sparse rewards. Unlike TIGER, ResFiT uses imitation as
the executable controller and learns residual corrections in action space;
TIGER uses imitation as a short-horizon task-space guide while allowing the RL
policy to learn its own behavior.

\noindent\textbf{Flow Matching:}
We use a flow-matching imitation policy~\citep{flow-matching,pi0} as the
behavior-cloning backbone. The policy learns a generative action-chunk
distribution by transporting noise to expert action sequences, providing an
expressive imitation prior for multimodal demonstrations. TIGER uses this
policy to generate action chunks for task-space guidance, and we also evaluate
the same policy directly as an imitation-only baseline. This isolates the
benefit of online RL with imitation-guided task-space rewards from the
performance of the underlying generative BC model.

\paragraph{ReWiND.}
ReWiND~\cite{rewind} learns a language-conditioned visual reward function ($R_\psi(o_{1},z)$) that maps a prefix of image observations ($o_{1}$) and a task instruction ($z$) to a dense, $[0,1]$-normalized estimate of task progress. In the original training recipe, the reward model is trained on successful language-labeled demonstrations together with a curated subset of Open-X. Matched sequence--instruction pairs receive increasing progress targets, mismatched pairs receive zero-progress targets, and video-rewind augmentation appends reversed frames to a forward trajectory prefix with decreasing progress targets along the reversed portion. Target-environment instructions are additionally augmented with LLM-generated alternatives. The original method uses this progress estimate, together with a success bonus, for offline policy pretraining followed by online RL fine-tuning.

\paragraph{Robometer.}
Robometer~\cite{liang2026robometer} is a general-purpose video--language reward model
that conditions on a task instruction and a trajectory video to predict dense, per-frame
task progress and task-success probabilities. It is trained with per-frame progress and
success losses together with a pairwise trajectory-preference loss. Preference pairs include
same-task comparisons between executions of different quality, different-task instruction negatives,
and synthetic failures produced by video rewinding, allowing the model to use expert demonstrations
as well as real and augmented suboptimal trajectories. We include Robometer as a complementary learned
reward baseline because its explicit pairwise preference objective differs from ReWiND's direct
progress-prediction objective and TIGER's task-space progress construction.

\subsubsection{Setup and Description}
\label{app:simulation_details}

We evaluate our approach on simulation benchmarks spanning four MetaWorld
tasks~\cite{metaworld} (Assembly, Box Close, Coffee Push, and Stick Pull),
three tasks from the Robomimic benchmark~\cite{robomimic2021}
(Can, Square, and ToolHang), and a humanoid
shape-sorting insertion task. Coffee Push is included as an additional
MetaWorld task in the appendix beyond the three MetaWorld tasks summarized in
the main paper. In the humanoid setting (Fig.~\ref{fig:gpr_benchmark}), the
robot must insert a cylinder into a shape sorter under tight tolerances and
substantial initial-state randomization ($35\text{cm} \times 35\text{cm}$ for
the cylinder and $4\text{cm} \times 4\text{cm}$ for the sorter), making
standalone imitation brittle.

All tasks use a sparse binary success reward except for the humanoid and
ToolHang environments, where we define intermediate sparse signals to support
curriculum learning. In the humanoid task, the environment provides two sparse
signals: lifting the cylinder to a target height and successfully inserting it
into the sorter. The final insertion reward remains difficult to discover from
random exploration, making the task challenging for standard sparse-reward RL.

In Robomimic ToolHang, the task has two stages: assembling the frame by
inserting the hook/frame into the stand, and then hanging the tool on the
assembled frame. We evaluate two sparse-reward settings. The first provides a
reward after the frame-insertion stage is completed, while the second provides a
reward only after full task completion, when the tool is successfully hanging on
the frame. The second setting follows the original Robomimic success condition,
which requires both frame assembly and tool hanging.

For ToolHang and humanoid shape sorting, we use curriculum learning shared
across all baselines. The curriculum is included to make the long-horizon sparse
tasks trainable within the fixed interaction budget, not as a TIGER-specific
advantage. For humanoid insertion, we use a five-stage curriculum that
progressively increases the randomization of the insertion setup. For ToolHang,
we use a two-level curriculum. In the first level, the behavior-cloning policy
executes the first 150 steps, bringing the hook near the stand but not
performing the insertion. Control is then transferred to the RL policy. Once a
method achieves 50\% success on the full task, including hook insertion and tool
hanging, we switch to the second level, where the RL policy controls the full
episode.

For the Robomimic tasks, we use two camera views per task. For MetaWorld and
humanoid shape sorting, we use a single camera view. In the humanoid task, we
use a one-dimensional action for hand posture. Across all tasks, all methods use
relative end-effector control.

We additionally conduct a targeted learned-reward comparison on MetaWorld Assembly,
Box Close, and Stick Pull and Robomimic Square. ReWiND, Robometer, and TIGER use the
same task-specific demonstrations, RLPD backend, interaction budget, and five-seed protocol.

We also evaluate on all ten LIBERO-Spatial tasks~\cite{liu2023libero}. These simulated tabletop tasks share the objective of placing a black bowl on a plate but vary the bowl's initial location or spatial relationship to the surrounding objects. We consider settings with 10 and 50 demonstrations per task and train one shared multi-task BC policy and one shared multi-task RL policy per method across all ten tasks, using a one-hot task identifier. Training and evaluation use the same task set, so this experiment measures multi-task learning rather than unseen-task generalization.

\subsubsection{Full Training Curves}
\label{app:sim_training_curves}
We provide the full training curves for all simulation tasks, including
per-task results, in this section. Table~\ref{tab:sim_summary} summarizes the
key milestones used to compare sample efficiency and final success. Each
experiment is evaluated over five random seeds.
\begin{figure}[!htbp]
  \centering
  \includegraphics[width=.45\linewidth]{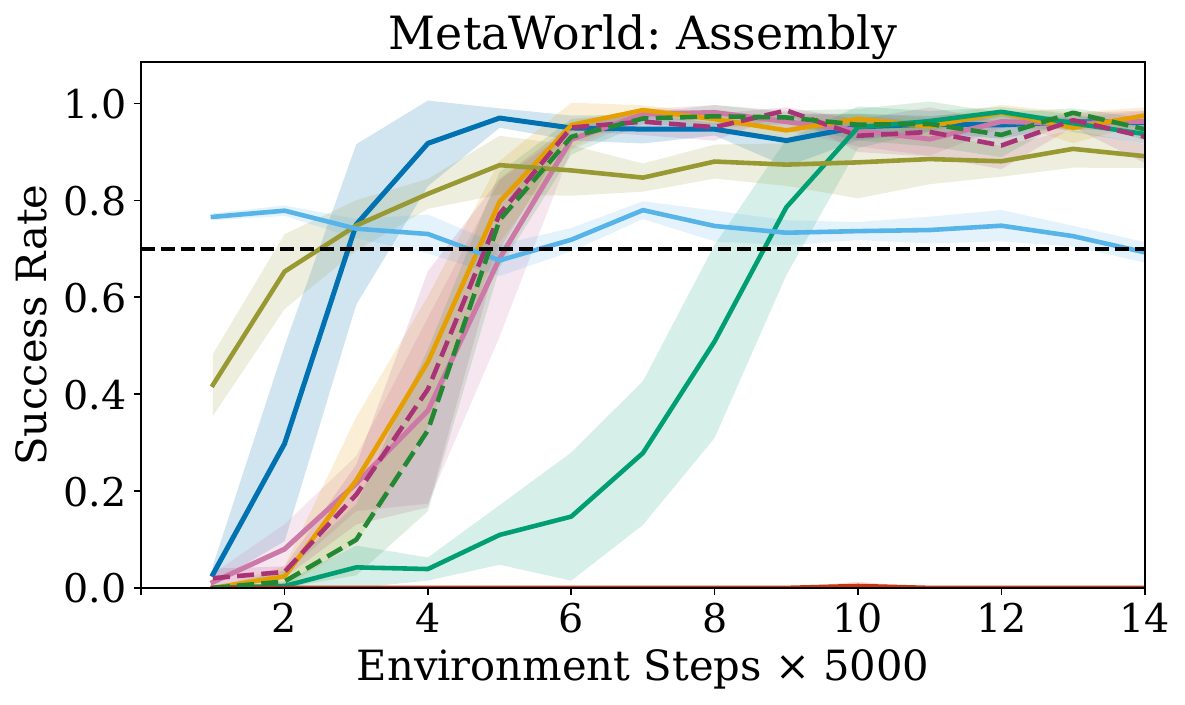}\hfill
  \includegraphics[width=.45\linewidth]{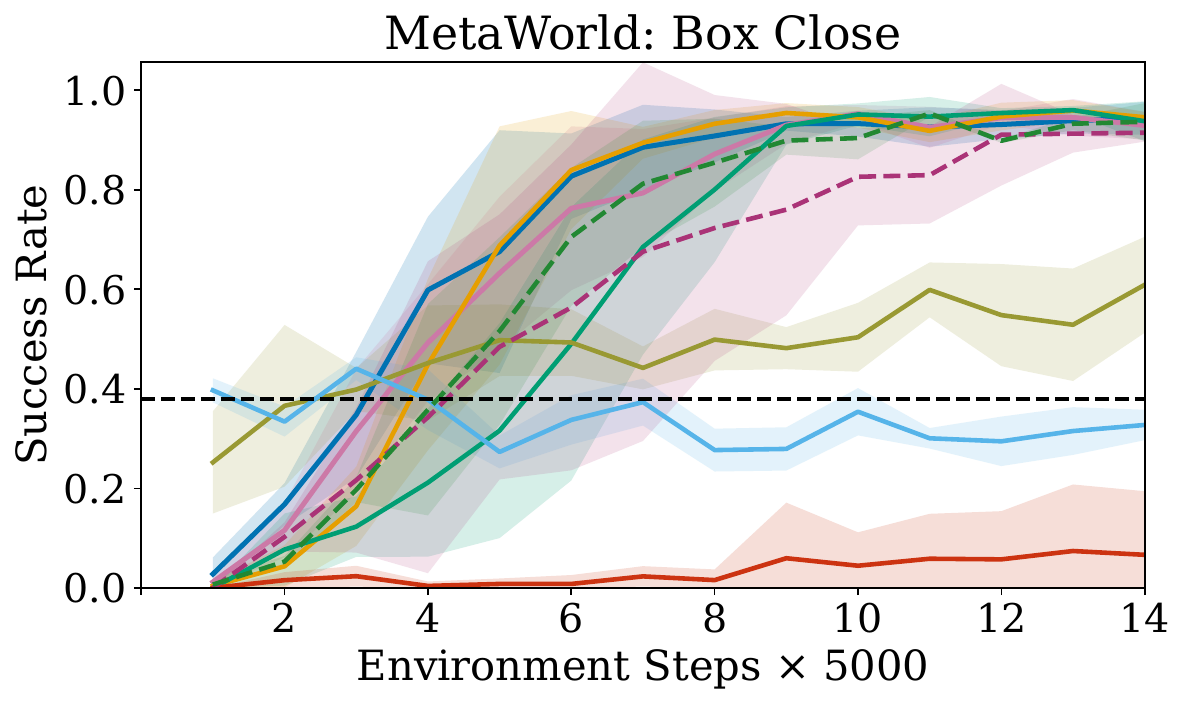}\\[0.2cm]
  \includegraphics[width=.45\linewidth]{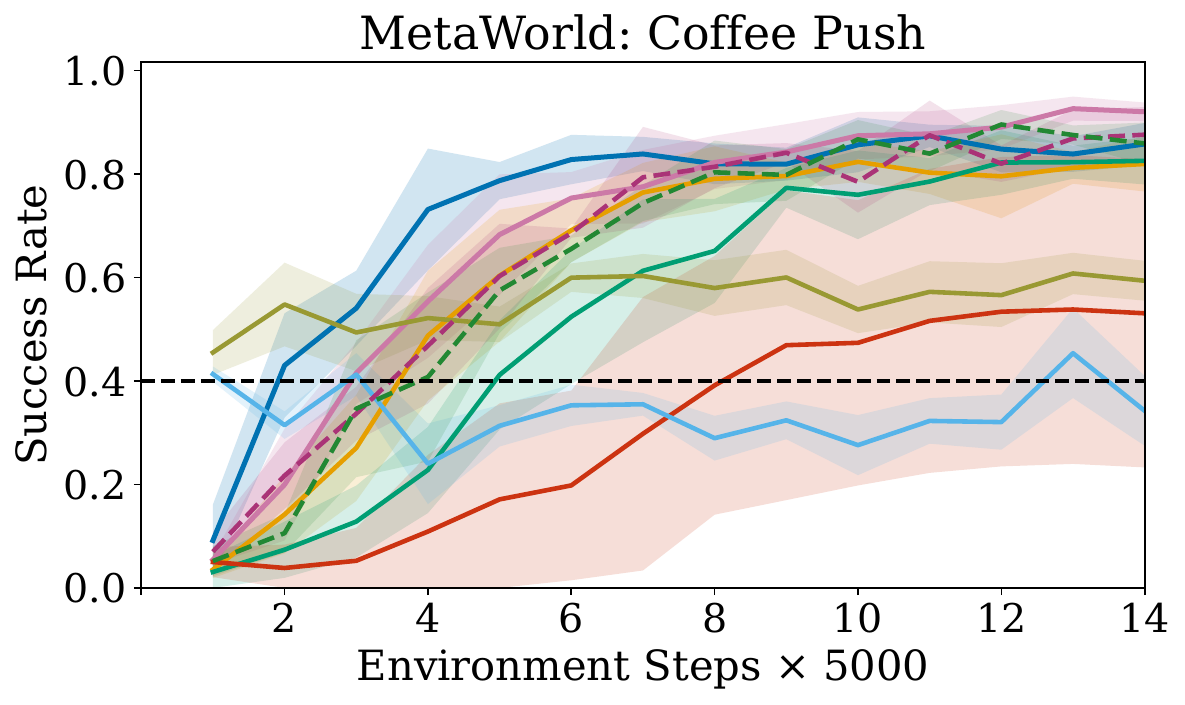}\hfill
  \includegraphics[width=.45\linewidth]{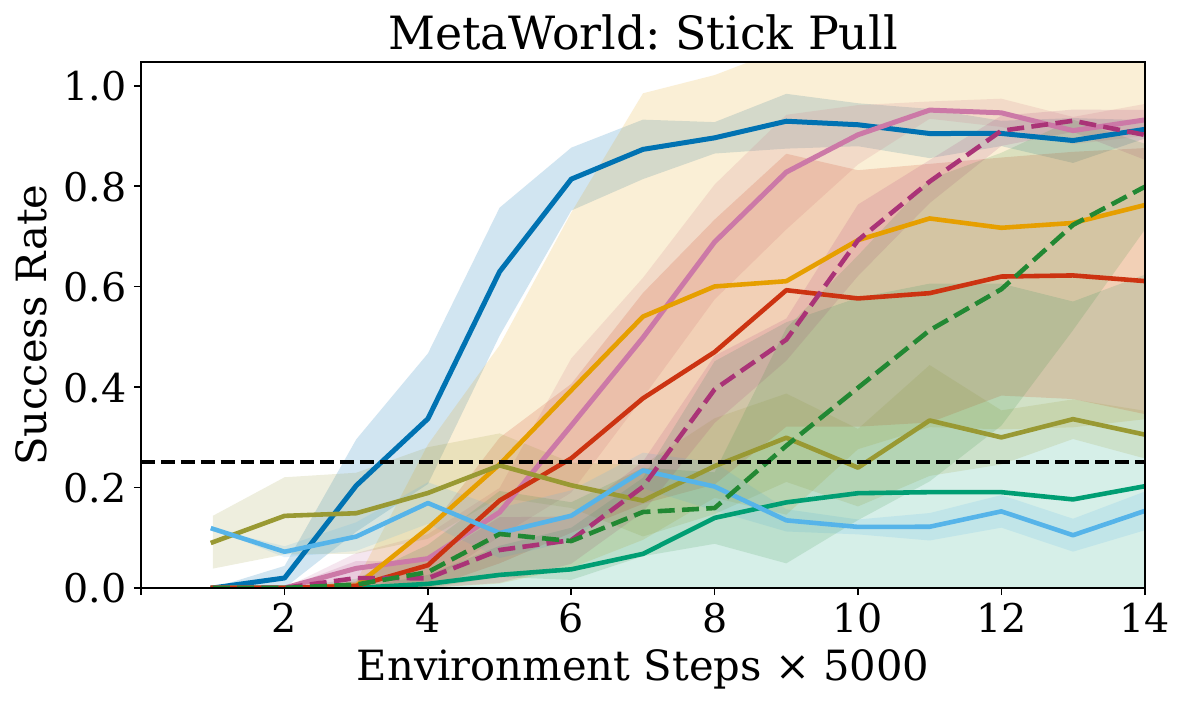}\\[-0.1cm]
  \includegraphics[width=.99\linewidth]{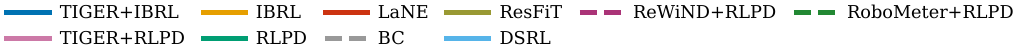}
  \caption{Performance on MetaWorld benchmark}
  \label{fig:meta_world}
\end{figure}

\begin{figure}[!htbp]
    \centering
    \includegraphics[width=.45\linewidth]{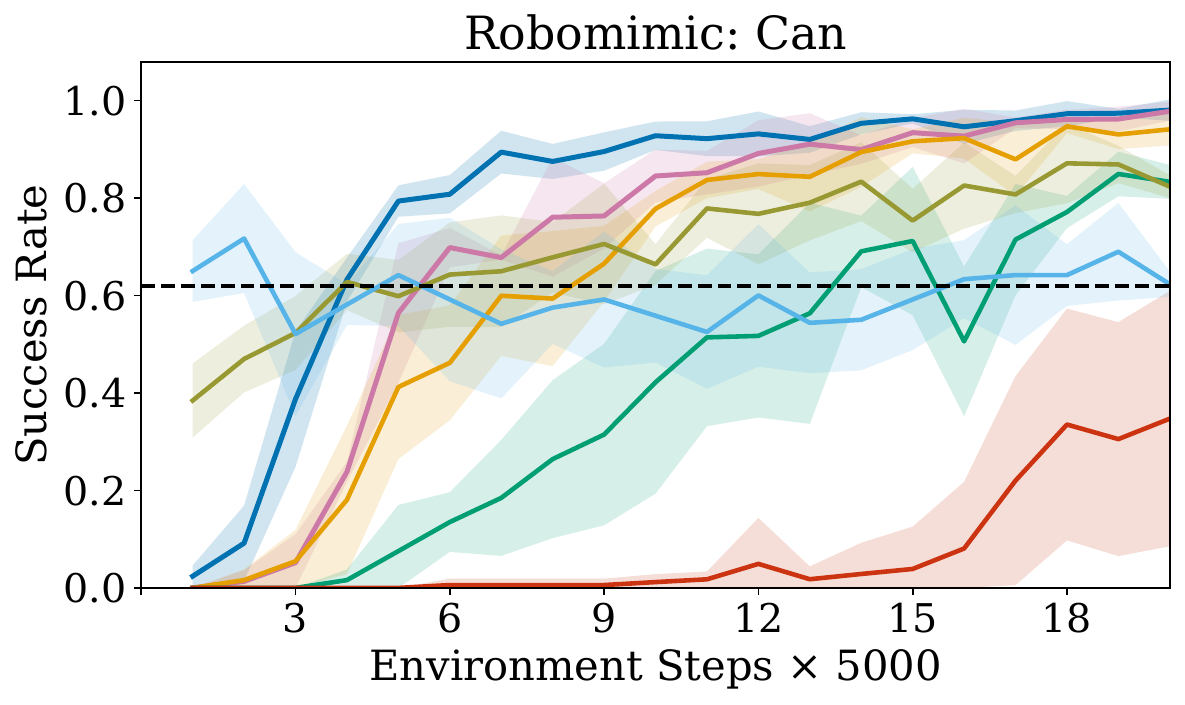}\hfill
    \includegraphics[width=.45\linewidth]{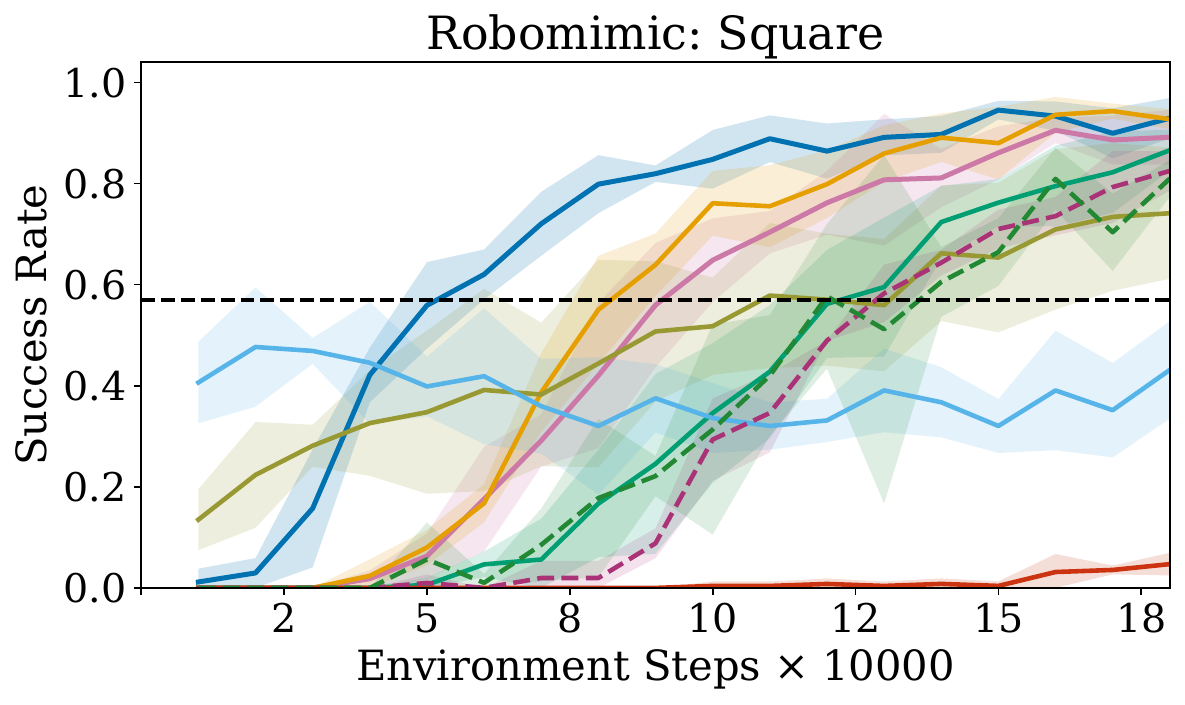}\\[0.2cm]
    \includegraphics[width=.45\linewidth]{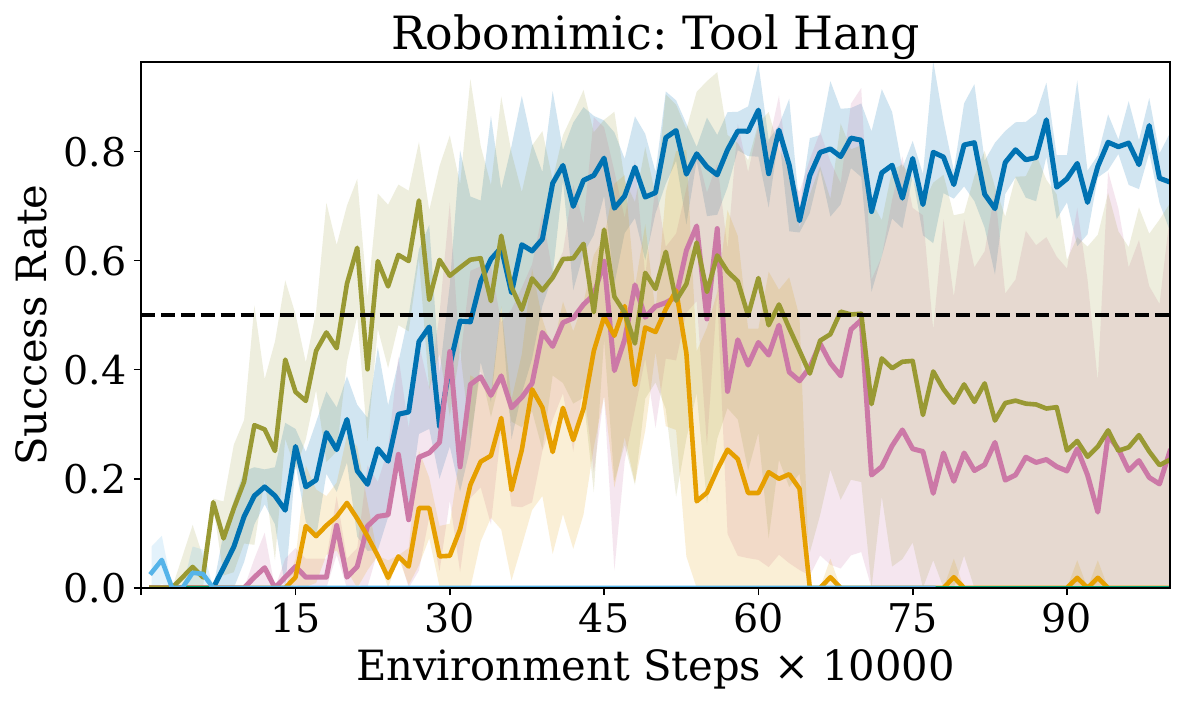}\hfill
    \raisebox{0.5\height}{\includegraphics[width=.40\linewidth]{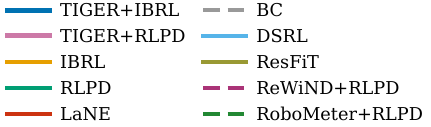}}
    \caption{Performance on Robomimic benchmark}
    \label{fig:robosuite}
\end{figure}

\begin{figure}[H]
    \centering
    \includegraphics[width=.40\linewidth]{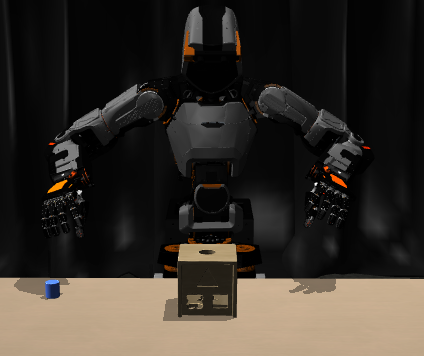}\hfill
    \includegraphics[width=.55\linewidth]{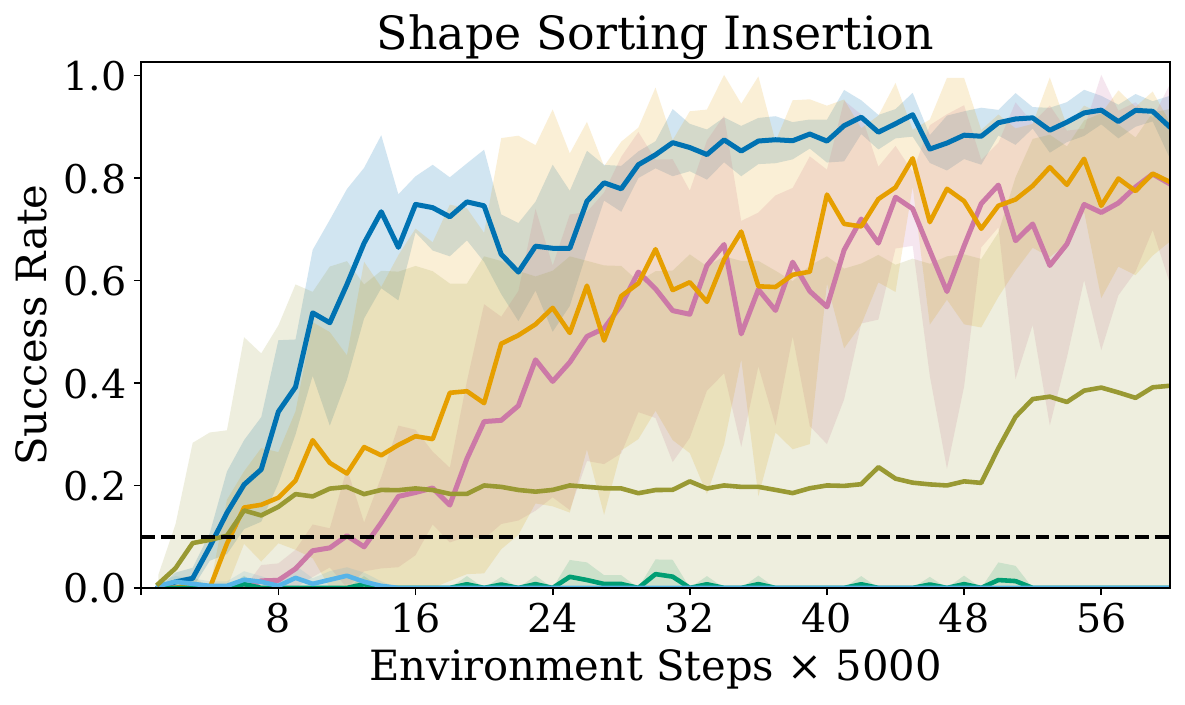}\\
    \includegraphics[width=.99\linewidth]{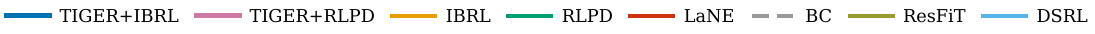}
    \caption{Performance on Humanoid Shape Sorting task}
    \label{fig:gpr_benchmark}
\end{figure}

\begin{figure}[H]
    \centering
    \includegraphics[width=.5\linewidth]{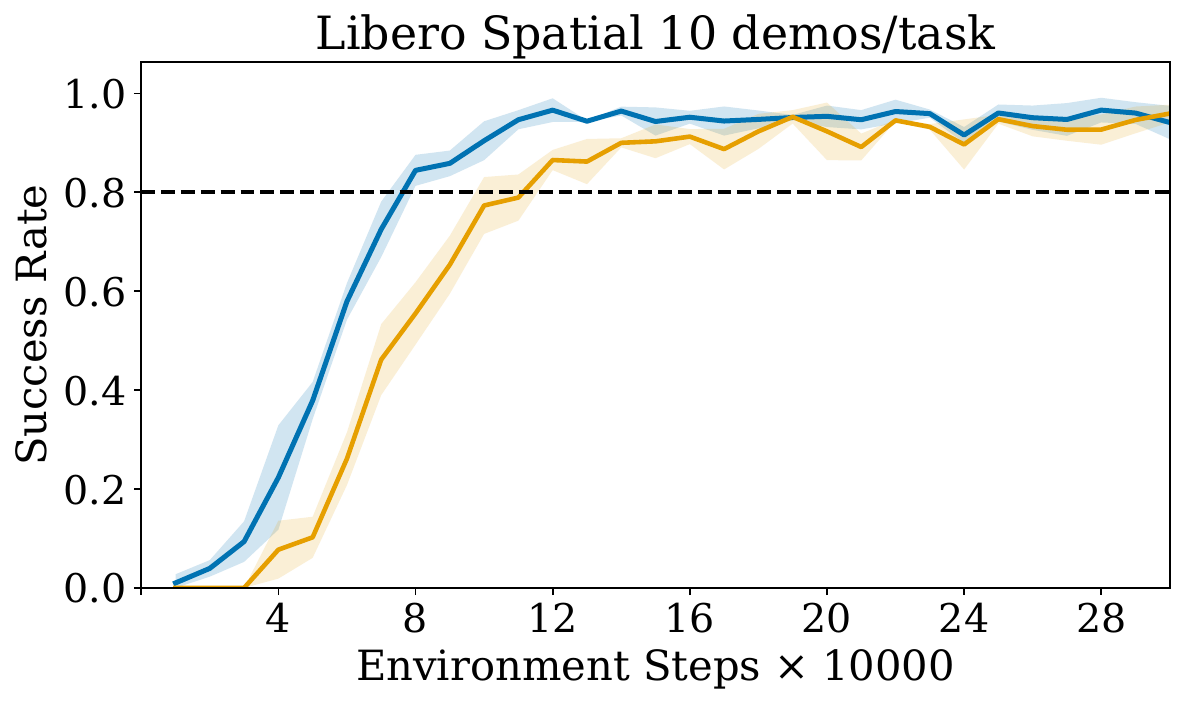}\hfill
    \includegraphics[width=.5\linewidth]{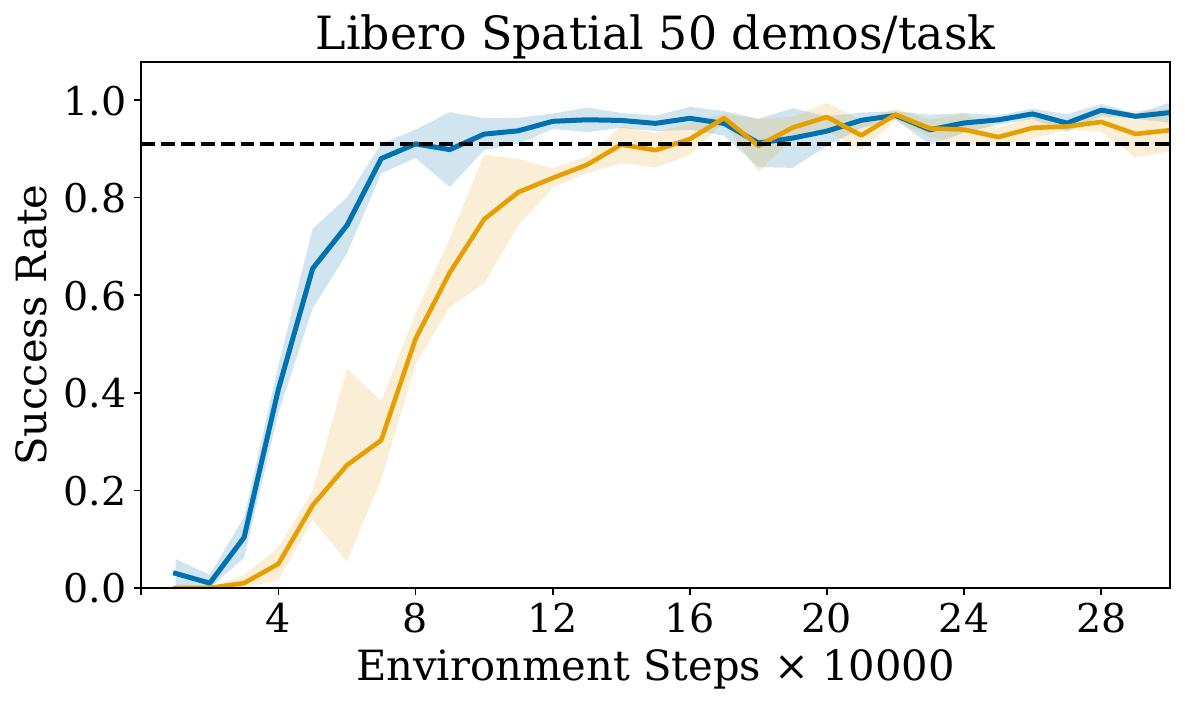}\\
    \includegraphics[width=.99\linewidth]{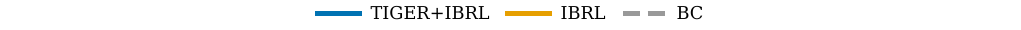}
    \caption{Multi-task LIBERO-Spatial learning curves.}
    \label{fig:libero_benchmark}
\end{figure}
\subsection{Real Robot Environment Details}
\label{app:real_detail}

This section details the real-robot task setups used in our experiments. A key
objective of the study is to evaluate learning across comparatively large
workspaces, where early-stage exploration can occasionally drive the robot into
unsafe or uninformative regions. Workspace limits are task-dependent and are
therefore enforced at the environment level to provide consistent training
signals and to explicitly measure safety violations, consistent with common
practice in real-world RL implementations. If the end-effector breaches these
limits for several consecutive steps, the episode is terminated and the
environment is reset, without applying any additional penalty reward. Commanded
actions are also down-scaled by the high-level controller before being
translated into low-level control signals.

\subsubsection{Block Picking}

\noindent\textbf{Description.}
The objective is to grasp a solid red block. At the start of each episode, the
block pose is uniformly sampled inside a \(25\text{ cm}\times30\text{ cm}\)
workspace that is fully visible to the wrist camera when the robot is at its
nominal home configuration. We collected ten human teleoperation demonstrations,
and each episode is capped at 100 environment steps. Controller action scales
are \((0.05, 0.20)\) for Cartesian position and orientation commands,
respectively.

\noindent\textbf{Operational workspace.}
Cartesian \emph{positions} are specified \emph{relative} to the initial
end-effector pose, whereas Cartesian \emph{orientations} (roll--pitch--yaw) are
\emph{absolute} in the world frame:
\begin{align*}
\mathbf{p}_{\text{low}}   &= [-0.19,-0.35,-0.90]\;\text{m},\\[4pt]
\mathbf{p}_{\text{high}}  &= [\,0.16,\;0.35,\;0.05\,]\;\text{m},\\[6pt]
\boldsymbol{\theta}_{\text{low}} &= [\,0.75\pi,0.0,0.0\,]\;\text{rad},\\[4pt]
\boldsymbol{\theta}_{\text{high}} &= [\,\pi,\;0.25\pi,\;\pi\,]\;\text{rad}.
\end{align*}
These bounds define the task-specific safety region used for training,
evaluation, and violation counting.

\noindent\textbf{Reset.}
After an episode terminates, the robot opens the gripper, releasing the object
if it was grasped, and returns to its home pose. The block is manually
repositioned within the defined workspace. During training, following
\cite{ibrl}, the block is initially placed near the center of the region; after
the robot achieves several successful grasps, the initial block pose is
gradually moved toward the workspace boundaries.

\noindent\textbf{Reward.}
The task uses an automated binary reward. Success is signaled only when all of
the following conditions are met: (i) the gripper is closed beyond a tightening
threshold, (ii) the object is held stably without slipping for at least three
consecutive time steps, and (iii) the object is lifted at least \(2\) cm above
its initial height.

\subsubsection{Drawer Opening}

\noindent\textbf{Description.}
In the \emph{Drawer Opening} task, the policy must pull open the second drawer
in a stack of plastic drawers. The drawers have small handles that are difficult
for the gripper to grasp; other drawers may be opened but yield no reward. The
task requires fine, smooth contact control and tests whether sparse-reward RL
can use demonstration-derived guidance to reliably engage the correct handle. We
collected ten teleoperation demonstrations. Controller action scales are
\((0.05, 0.20)\) for Cartesian position and orientation commands, respectively,
and each episode is capped at 100 steps.

\noindent\textbf{Operational workspace.}
Cartesian \emph{positions} are specified \emph{relative} to the initial
end-effector pose, whereas Cartesian \emph{orientations} (roll--pitch--yaw) are
\emph{absolute} in the world frame:
\begin{align*}
\mathbf{p}_{\text{low}}   &= [-0.06,-0.10,-0.21]\;\text{m},\\
\mathbf{p}_{\text{high}}  &= [\,0.10,\;0.12,\;0.20\,]\;\text{m},\\[4pt]
\boldsymbol{\theta}_{\text{low}} &= [\,0.45\pi,\;0,\;0\,]\;\text{rad},\\
\boldsymbol{\theta}_{\text{high}} &= [\,\pi,\;0.20\pi,\;0.20\pi\,]\;\text{rad}.
\end{align*}
These bounds define the task-specific safety region used for training,
evaluation, and violation counting.

\noindent\textbf{Reset.}
After an episode terminates, the robot opens the gripper if necessary, returns
to its home pose, and the opened drawer is manually closed. The next episode
begins with \(15\%\) joint-space pose randomization.

\noindent\textbf{Reward.}
A binary reward is issued once the second drawer, marked with red tape, is
detected as fully open by an auxiliary reward camera. Partial openings or
openings of other drawers do \emph{not} yield reward.

\subsubsection{Rotary Insertion}

\noindent\textbf{Description.}
The \emph{Rotary Insertion} task requires the robot to pick up a blue object and
insert it into a designated hole marked in white. The insertion point can rotate
around the vertical axis, and the object's initial position is randomized within
a \(15\text{ cm}\times15\text{ cm}\) region. The task is challenging because it
combines long-horizon grasping, spatial search, target tracking, and precise
contact-rich insertion. We collected 50 teleoperation demonstrations.
Controller action scales are \((0.05, 1.5)\) for Cartesian position and
orientation commands, respectively, and each episode is capped at 150 steps.

To improve learning stability, we use curriculum learning during training. In
each curriculum stage, the disk orientation is advanced by \(45^\circ\), and
the policy must successfully complete the task five times before progressing.
Previous stages are sampled non-uniformly to mitigate catastrophic forgetting.
After completing the full rotation cycle, we randomize the insertion location;
in the final stage, the policy must insert the object while the disk is actively
rotating.

\noindent\textbf{Operational workspace.}
Cartesian \emph{positions} are specified \emph{relative} to the initial
end-effector pose, whereas Cartesian \emph{orientations} (roll--pitch--yaw) are
\emph{absolute} in the world frame:
\begin{align*}
\mathbf{p}_{\text{low}} &= [-0.25,-0.35,-0.10]\;\text{m}, \\
\mathbf{p}_{\text{high}} &= [\,0.15,\;0.35,\;0.50\,]\;\text{m}, \\[6pt]
\boldsymbol{\theta}_{\text{low}} &= [\,0.9\pi,-0.1\pi,-0.4\pi\,]\;\text{rad}, \\
\boldsymbol{\theta}_{\text{high}} &= [\,\pi,\;0.1\pi,\;0.4\pi\,]\;\text{rad}.
\end{align*}
These bounds define the task-specific safety region used for training,
evaluation, and violation counting.

\noindent\textbf{Reset.}
At episode termination, the robot opens the gripper, releasing the object if it
was grasped, and returns to the initial end-effector pose. The object is then
placed at a random position following the same center-out curriculum used in
Block Picking.

\noindent\textbf{Reward.}
A binary reward is given once an infrared sensor on the rotating device detects
that the object is fully seated in the insertion hole; the episode then
terminates immediately.

\subsection{Extended Ablation Study}
\label{app:extended_ablation}
We provide additional ablations that isolate the main design choices in TIGER:
the imitation-guided reward and target observation, the quality of the
demonstrations used to train the imitation policy, reward-scaling sensitivity,
the offline pretraining objective, and the transferability of the
controller-aware mapping. Since TIGER+IBRL achieves the strongest overall
performance in both simulation and real-robot experiments, we use IBRL as the
default RL backend in this analysis unless otherwise stated.

\noindent\textbf{Effect of imitation-guidance design.}
To isolate the contribution of each guidance channel, we compare full TIGER
against two variants: one that uses the imitation-guided dense reward but does
not provide the target pose as an observation, and one that provides the target
pose as an observation but removes the dense reward. Results are shown in
Fig.~\ref{fig:ablation_rew_obs}.

Both components improve learning relative to their ablated counterparts.
Removing either the dense reward or the target-pose observation reduces sample
efficiency, indicating that reward shaping and goal conditioning provide
complementary guidance. The dense-reward-only variant outperforms the
target-observation-only variant, suggesting that the imitation-guided reward is
the primary driver of online learning, while the target observation helps the
policy condition its actions on the local reference.
\begin{figure}[H]
  \centering
  \includegraphics[width=.45\linewidth]{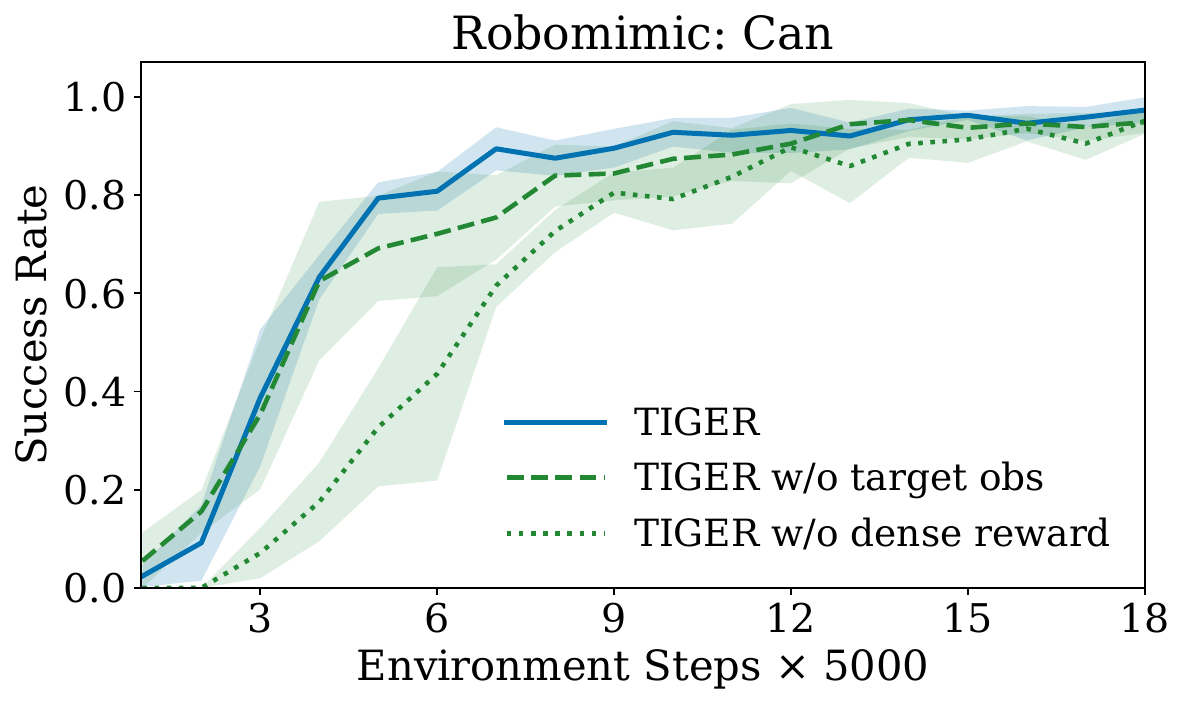}\hfill
  \includegraphics[width=.45\linewidth]{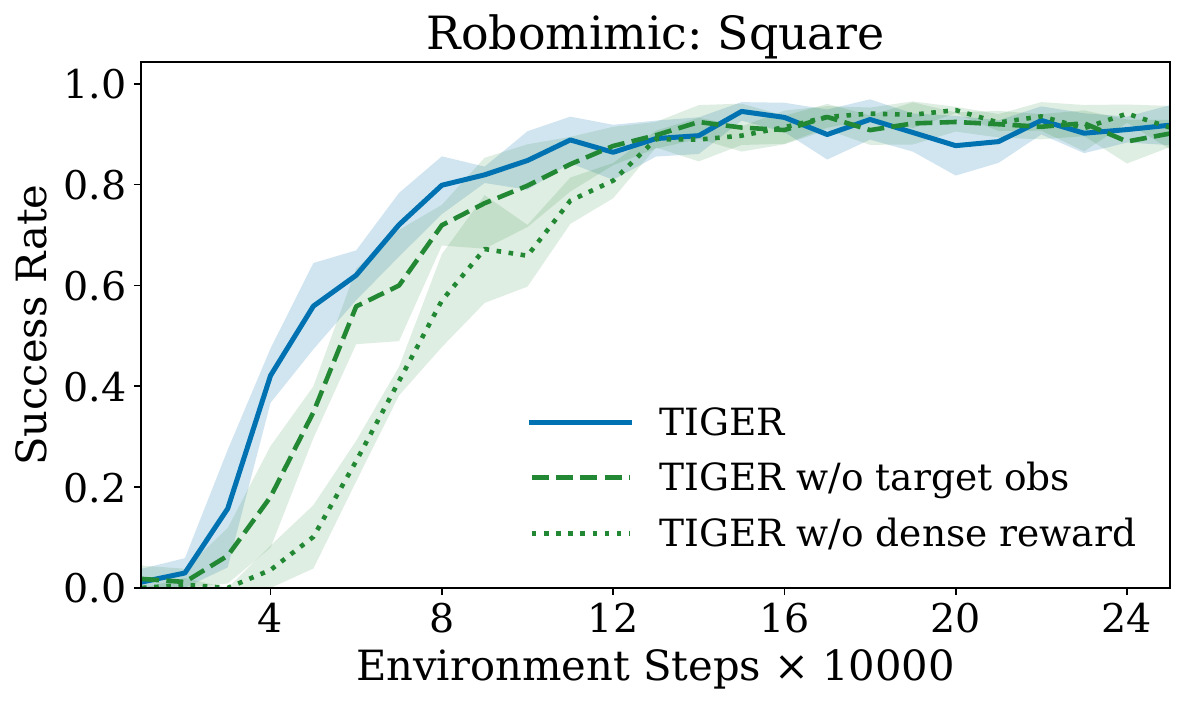}\\[0.1cm]
  \caption{Effect of imitation-guidance design.}
  \label{fig:ablation_rew_obs}
\end{figure}

\noindent\textbf{Effect of demonstration quality.}
To evaluate sensitivity to demonstration quality and variability, we compare
TIGER and IBRL on Robomimic Can and Square using proficient-human (PH) and
multi-human (MH) demonstrations. PH data are collected from a single proficient
demonstrator and contain faster, more consistent trajectories, especially in
grasp pose. MH data are collected from multiple demonstrators and contain more
diverse, slower, and less consistent behaviors.

For each data source, we use 20 demonstrations for Can and 100 demonstrations
for Square. We train the same flow-matching BC policy used by both TIGER and
IBRL, and report both learning curves and final average episode length as a
proxy for policy efficiency. Figure~\ref{fig:data_quality_ablation} summarizes
the results.

Across both tasks and data sources, TIGER improves sample efficiency over the
matched IBRL baseline. The gains are largest when the imitation policy provides
locally consistent guidance, but TIGER is not limited to reproducing the BC
policy. Even with more variable MH demonstrations, the imitation-guided signal
provides useful early exploration structure, while online RL continues to
optimize the original sparse task reward. The episode-length results in
Figure~\ref{fig:data_quality_ablation}-bottom show that TIGER can recover more
efficient task solutions than the demonstrations/BC policy while maintaining
high final success.

The effect of demonstration quality is task-dependent. On Can, TIGER trained
with MH demonstrations remains more sample efficient than IBRL trained with the
same data. On Square, TIGER with MH data is slightly less sample efficient than
with PH data, consistent with the larger gap between the corresponding BC
policies and the greater variability of grasp and insertion behaviors in MH
demonstrations. Nevertheless, TIGER still recovers strong final performance,
indicating that imperfect imitation guidance mainly affects early exploration
rather than preventing online RL improvement.

\begin{figure}[H]
  \centering
  \includegraphics[width=.45\linewidth]{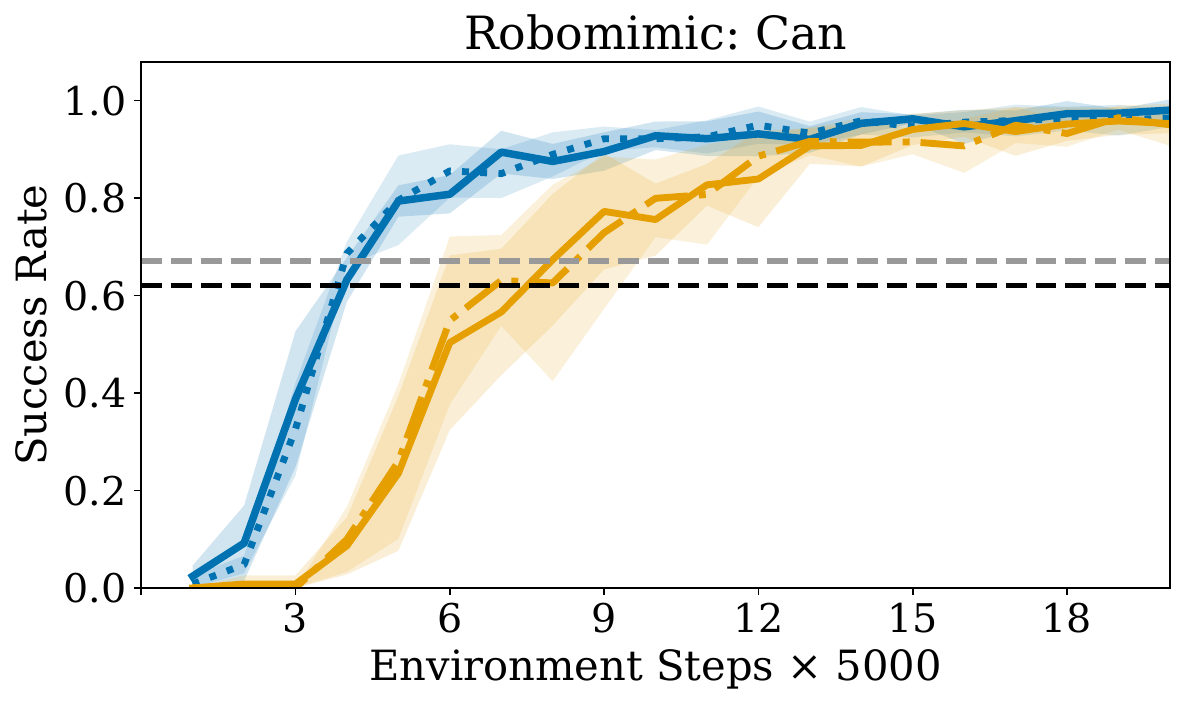}\hfill
  \includegraphics[width=.45\linewidth]{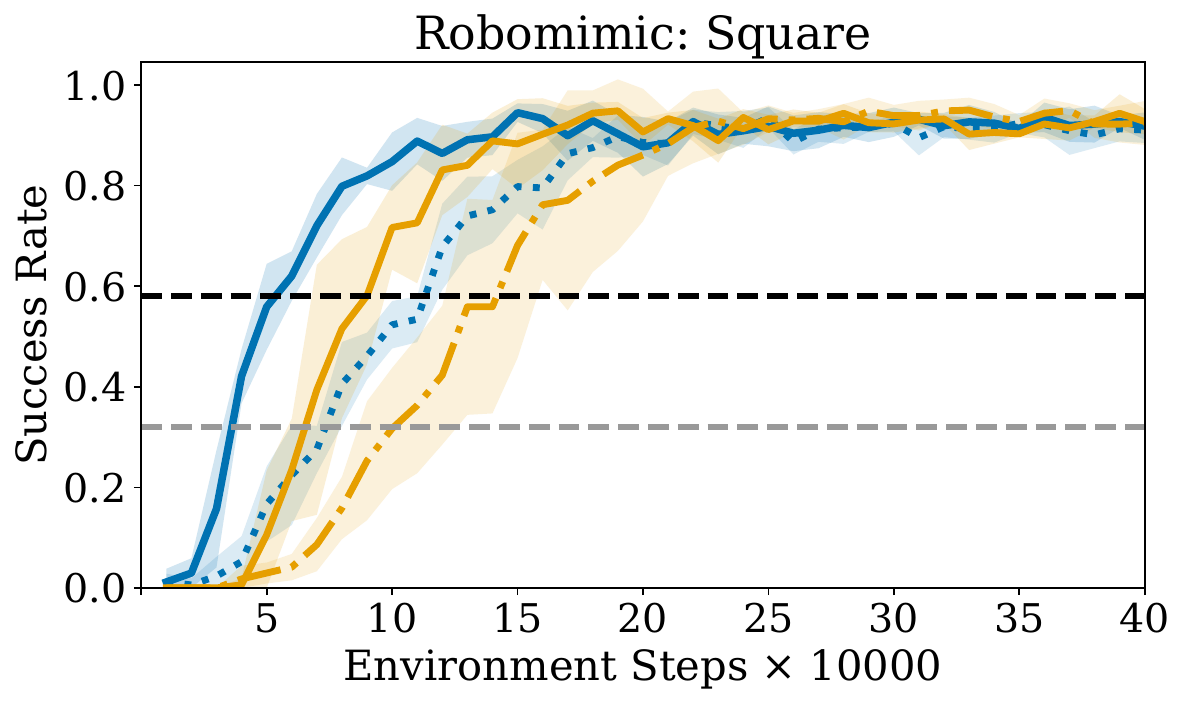}\\[0.1cm]
  {\centering\includegraphics[width=.7\linewidth]{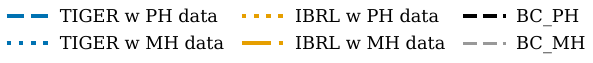}\par}\vspace{0.2cm}
  \includegraphics[width=.45\linewidth]{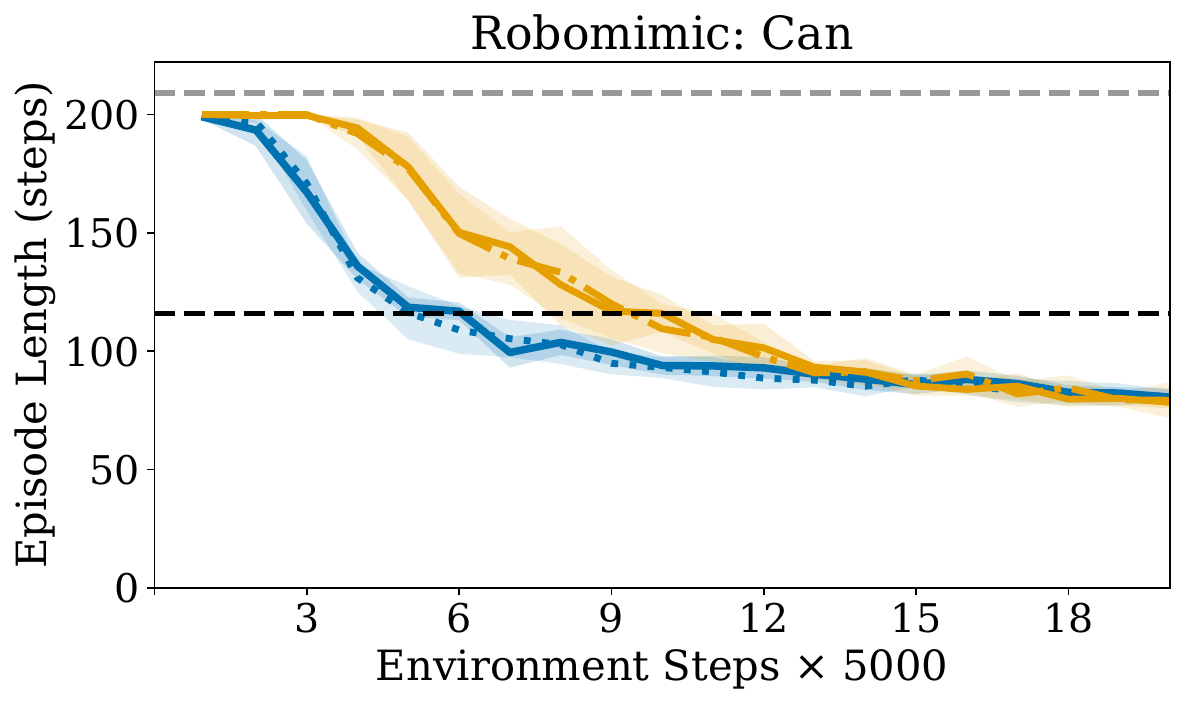}\hfill
  \includegraphics[width=.45\linewidth]{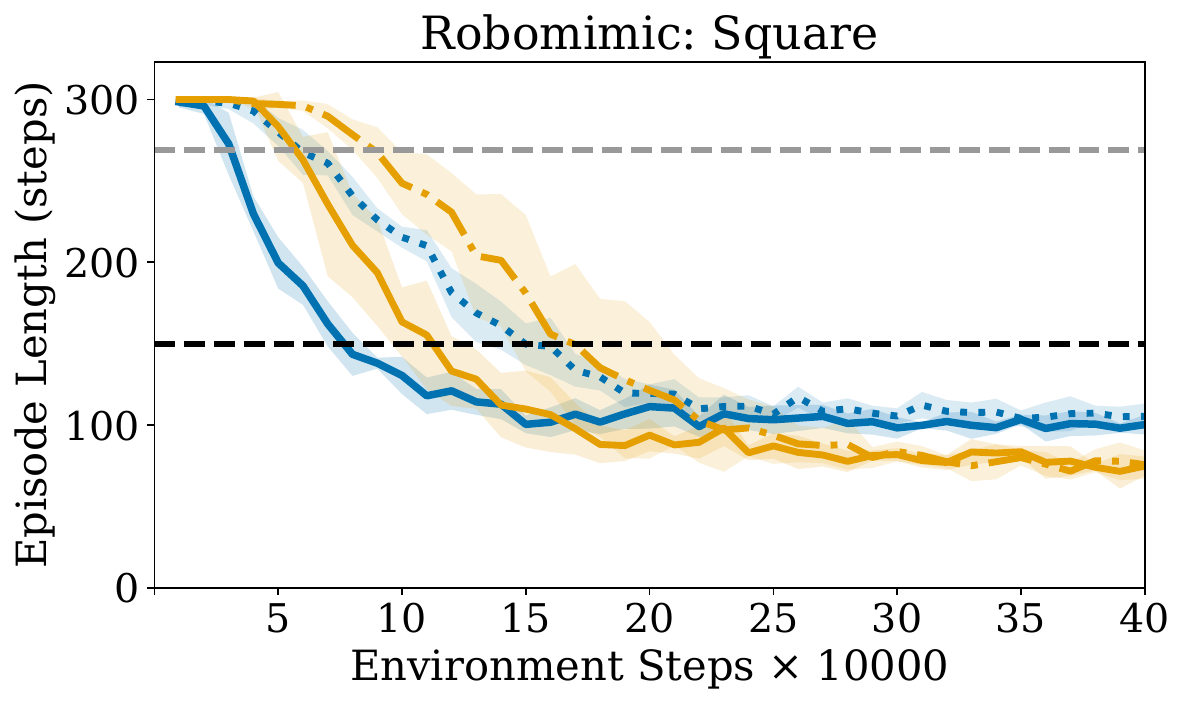}\\[0.1cm]
  {\centering\includegraphics[width=.7\linewidth]{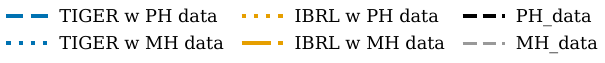}\par}
  \caption{Effect of demonstration quality on learning performance}
  \label{fig:data_quality_ablation}
\end{figure}
\noindent\textbf{Sensitivity to reward scaling.}
\label{app:ablation_k}
We study the effect of the reward-scaling factor $k$
(Eq.~\ref{eq:igrl_return}) on Can and Square.
This factor controls the relative scale between the sparse environment reward
and the cumulative imitation-guided shaping return. As shown in
Fig.~\ref{fig:ablation_k}, TIGER is robust to moderate changes in this scale:
all settings with $k \geq 1$ achieve similar convergence speed and final
success. Performance is slightly weaker for $k=0.5$, where imitation shaping can
become too large relative to the task reward. These results support the design
choice of scaling the sparse success reward to dominate the total shaping return
of an unsuccessful episode.
\begin{figure}[H]
    \centering
    \includegraphics[width=0.45\linewidth]{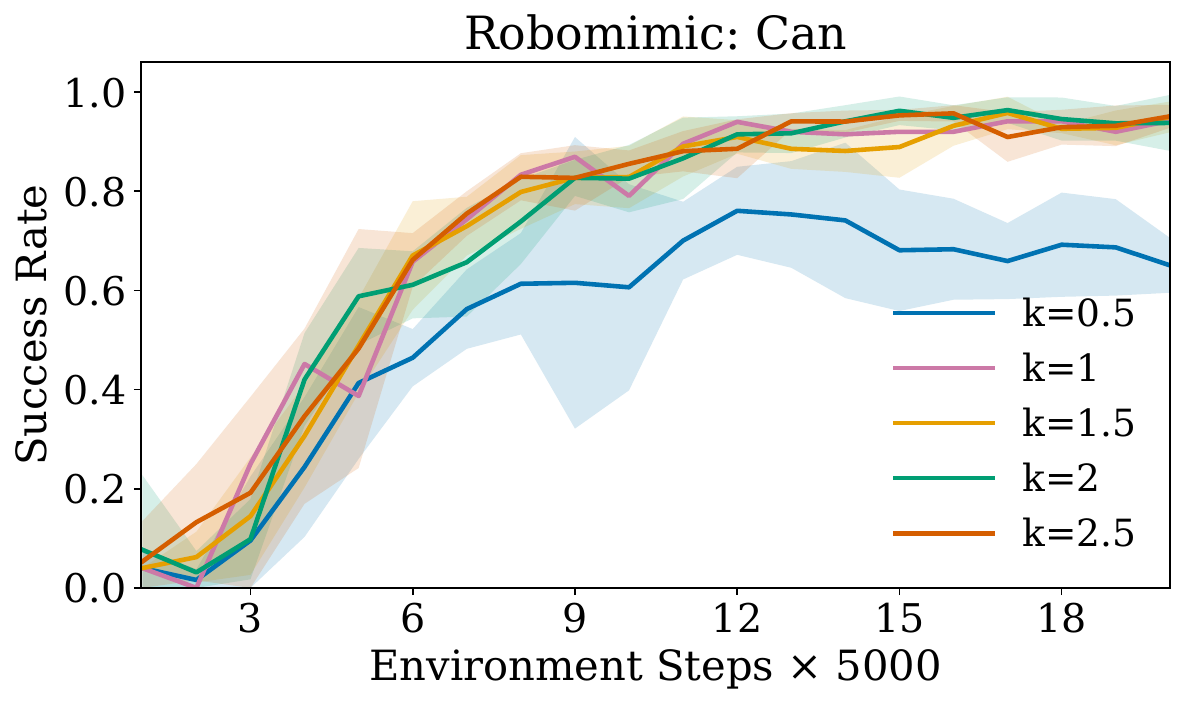}
    \includegraphics[width=0.45\linewidth]{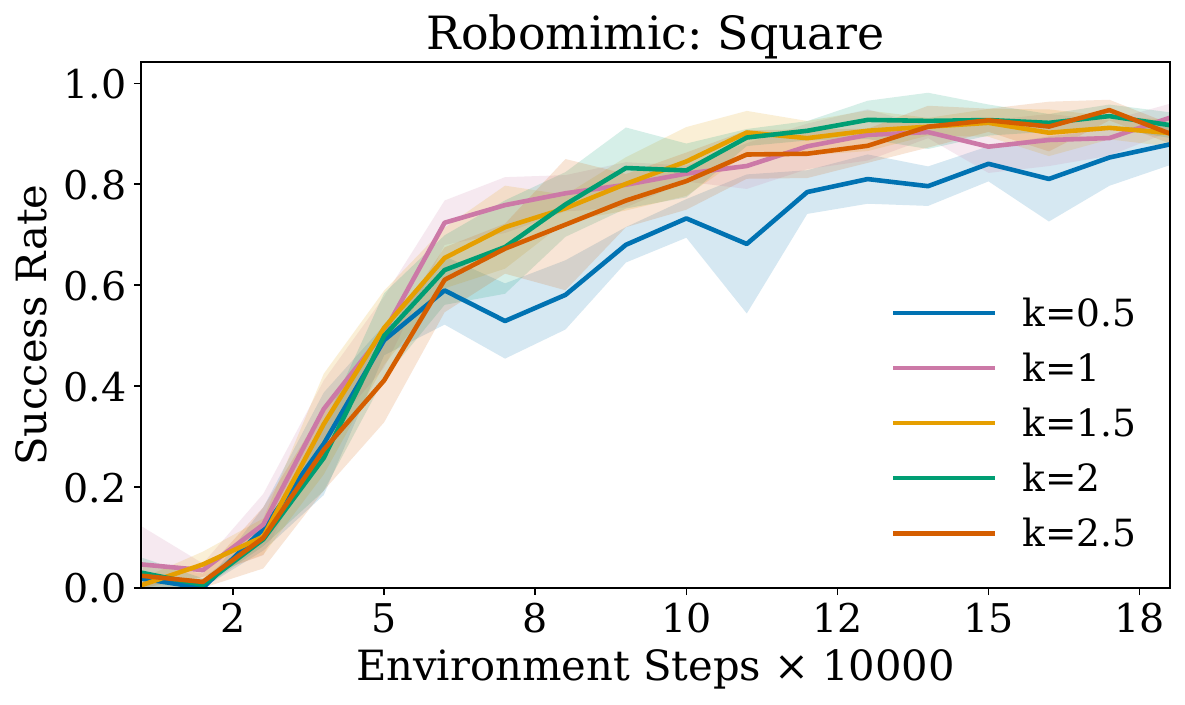}
    \vspace{-0.18cm}
    \caption{Effect of the reward-scaling factor $k$ on success rate.}
    \label{fig:ablation_k}
\end{figure}

\noindent\textbf{Pretraining objective on Robomimic Can.}
We additionally repeat the pretraining-objective ablation on Can to
verify that the trend observed on Square is not task-specific. As
shown in Fig.~\ref{fig:offline_robosuite_can}, the full TIGER pretraining
objective produces stronger offline policies than CalQL with sparse rewards or
CalQL augmented only with the TIGER dense reward. This supports the conclusion
that the benefit of pretraining comes not only from adding dense rewards, but
from using imitation-guided look-ahead to shape the conservative value
regularizer.
\begin{figure}[H]
  \centering
  \includegraphics[width=0.45\linewidth]{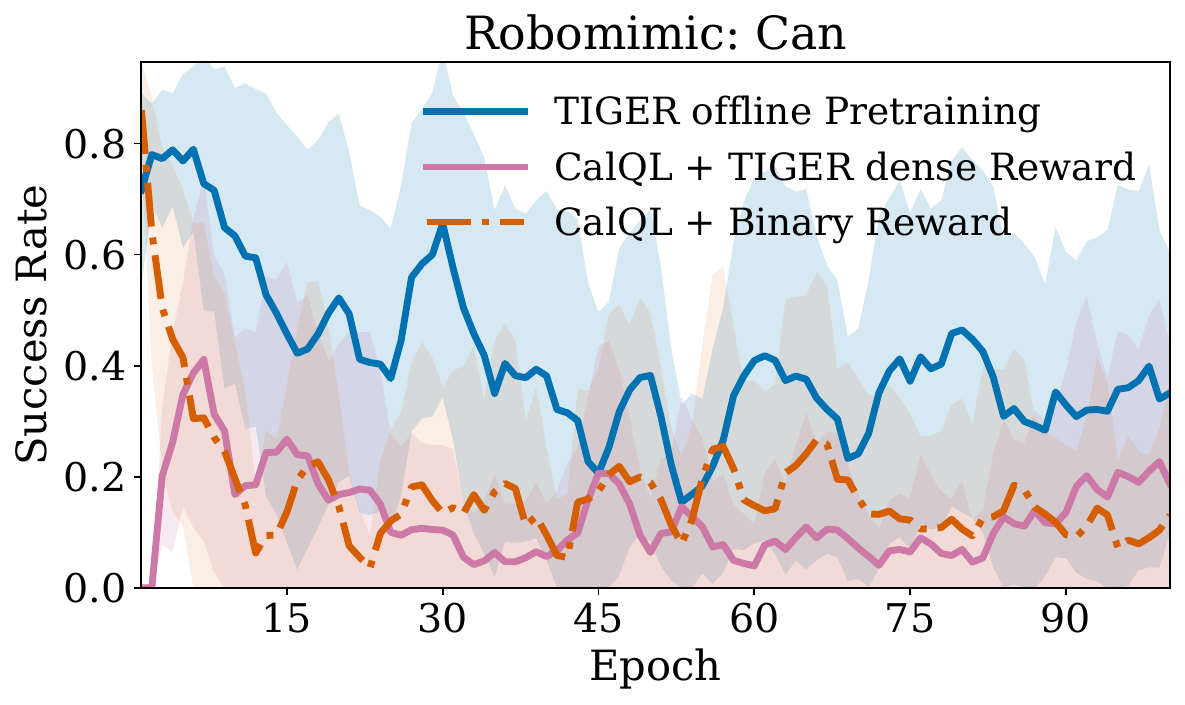}
  \caption{Pretraining objective ablation on Robomimic Can.}
  \label{fig:offline_robosuite_can}
\end{figure}

\noindent\textbf{Transferability of the controller-aware mapping.}
The controller-aware trajectory mapping
(Sec.~\ref{subsec:controller_aware_trajectory_mapping}, App.~\ref{app:mapping})
is fit from demonstrations on a single task for each robot/controller pair. To
test whether the learned correction is task-specific, we fit all candidate
mappings---identity, affine-only, FIR-only ARX, and full ARX---on Robomimic
Square using 160 training and 40 validation demonstrations, and evaluate the
fixed mappings on the full Robomimic Can dataset, containing 200 unseen
demonstrations collected with the same Franka Panda arm and OSC\_POSE
controller.

Figure~\ref{fig:mapping-transfer} reports per-step position-prediction error
across five motion phases: \emph{free-space} motion without contact,
\emph{pre-grasp} approach, \emph{grasp-event} finger--object contact,
\emph{carrying} after grasp closure, and \emph{contact-stalled} motion where
tracking is damped by environmental contact, such as insertion, bin-wall
contact, or table contact.
\begin{figure}[H]
  \centering
  \includegraphics[width=.99\linewidth]{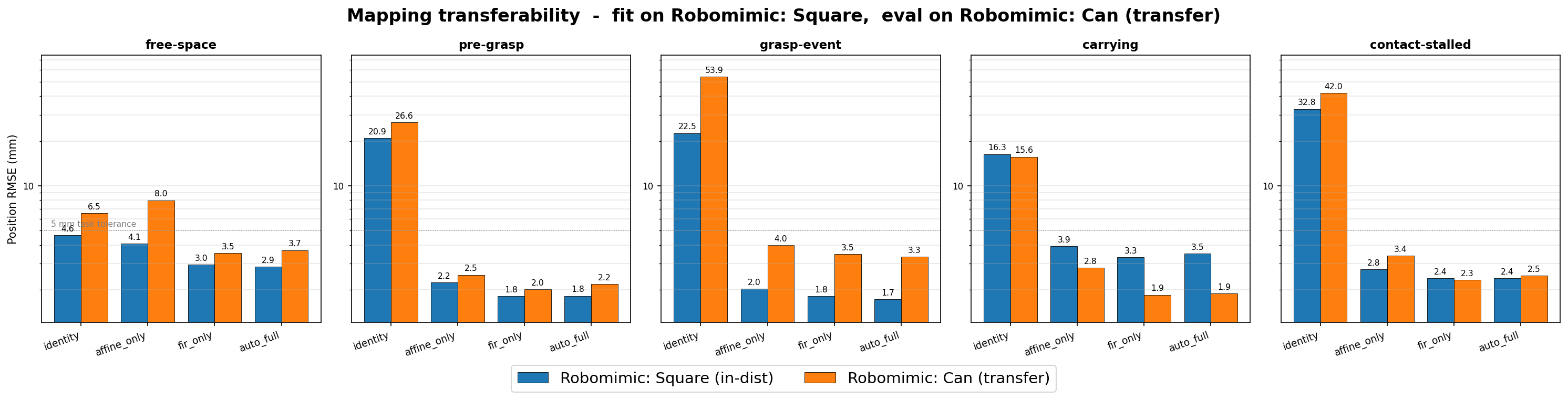}
  \caption{Transferability of the controller-aware mapping.}
  \label{fig:mapping-transfer}
\end{figure}

For the best-performing variant, FIR-only ARX, the contact-stalled error changes
from $2.39\,\mathrm{mm}$ in distribution to $2.34\,\mathrm{mm}$ under transfer,
a shift of only $0.05\,\mathrm{mm}$. Across free-space, pre-grasp, and
grasp-event phases, transfer error remains within $1.7\,\mathrm{mm}$ of the
in-distribution baseline, and carrying error is lower on Can than on Square
($1.85$ vs.~$3.31\,\mathrm{mm}$). In contrast, the identity baseline degrades
from $32.8\,\mathrm{mm}$ to $42.0\,\mathrm{mm}$ in the contact-stalled phase.
These results indicate that the learned correction primarily captures
controller-dependent execution effects, including latency, attenuation, and
finite tracking bandwidth, rather than task-specific motion patterns. Thus, a
single mapping can be reused across tasks on the same robot/controller pair.
\section{Controller-Aware Trajectory Mapping}
\label{app:mapping}

This appendix expands the three-stage trajectory-mapping pipeline introduced in
Sec.~\ref{subsec:controller_aware_trajectory_mapping}. Throughout, let
$a_t \in [-1,1]^{d_a}$ denote the normalized action representation predicted by
the IL policy at time $t$, and let $u_t \in \mathbb{R}^{d_u}$ denote the
corresponding command in the controller output space. We write the realized
end-effector pose as $x_t=(p_t,q_t)$, where $p_t\in\mathbb{R}^3$ is position
and $q_t\in\mathbb{H}$ is a unit quaternion. The realized per-step
end-effector displacement is represented as
$\Delta x_t=(\Delta p_t,\omega_t)$, with
$\Delta p_t=p_{t+1}-p_t$ and $\omega_t\in\mathbb{R}^3$ the axis-angle
orientation increment. The corresponding quaternion increment is
$\Delta q_t^{\mathrm{quat}}=\mathrm{Exp}(\omega_t)$, where
$\mathrm{Exp}$ denotes the $SO(3)$ exponential map.

\subsection{Scope and Limitations}
\label{app:mapping-scope}

The mapping models only low-order, time-invariant action-to-motion effects that
arise from controller preprocessing and finite execution bandwidth. It is
\emph{not} intended to predict contact transitions, force interactions, object
motion, or task-specific physics. We use a linear model intentionally because
(i) it admits a closed-form fit that is robust at the demonstration sample
sizes typical for sparse-reward manipulation (5--50 demos);
(ii) the mapping is used only to construct local guidance, while online RL
optimizes the true environment reward; and (iii) the dense reward in
Eq.~\eqref{eq:igrl_reward} only requires the reference trajectory to be
\emph{directionally} correct, not pose-accurate to millimeters. In contact-rich
phases, the model should therefore be interpreted as correcting systematic
controller-induced attenuation or delay in the commanded end-effector motion,
not as predicting the outcome of the contact interaction itself.

\subsection{Stage 1: Known Controller Transformations}
\label{app:mapping-known}

The first stage applies analytic transformations between the normalized action
representation used by the IL policy and the command space consumed by the
low-level controller.

\noindent\textbf{(a) Per-axis action rescaling.}
Given controller command bounds $u_{\min},u_{\max}\in\mathbb{R}^{d_u}$, we map
normalized actions to controller-space commands by
\begin{equation}
  u_t = \mathrm{diag}(s_u)\, a_t + b_u,
  \qquad
  s_u = \tfrac{1}{2}(u_{\max} - u_{\min}),
  \quad
  b_u = \tfrac{1}{2}(u_{\max} + u_{\min}).
\end{equation}
For the Robomimic Franka Panda OSC\_POSE controller,
$s_u=(0.05,0.05,0.05,0.5,0.5,0.5)$ in $(\mathrm{m},\mathrm{rad})$, so a
unit normalized command along one translational axis corresponds to $5$ cm and
a unit rotational command corresponds to approximately $28.6^\circ$.

\noindent\textbf{(b) Frame correction for orientation.}
The IL policy and controller may express axis-angle commands in different
reference frames. We therefore apply a fixed rotation $R_{\mathrm{fix}}$ to the
orientation component of $u_t$ before the residual mapping. In our setup,
$R_{\mathrm{fix}}$ is a $-90^\circ$ rotation about the world $z$-axis, aligning
the IL action convention with the controller convention.

\subsection{Stage 2: Learned Residual Mapping}
\label{app:mapping-learned}

Even after Stage~1, the realized end-effector displacement can differ
systematically from the controller-space command $u_t$ due to latency,
finite-bandwidth tracking, and controller-induced attenuation. We model this
action-to-motion mismatch with a linear map fit from demonstration pairs
$(u_t,\Delta x_t)$, where $\Delta x_t=(\Delta p_t,\omega_t)\in\mathbb{R}^6$.
We consider three model families and select among them using held-out
validation MSE.

\noindent\textbf{Family I: affine map with temporal delay.}
The simplest model assumes the realized displacement is an affine function of a
single delayed command:
\begin{equation}
  \Delta\hat{x}_t = W\, u_{t-\delta} + b,
\end{equation}
where $W\in\mathbb{R}^{6\times d_u}$ and $b\in\mathbb{R}^6$. The parameters
are estimated by ridge regression,
\begin{equation}
  (W^\star, b^\star)
  = \arg\min_{W, b}
    \frac{1}{|\mathcal{T}_\delta|}
    \sum_{t\in\mathcal{T}_\delta}
      \bigl\| \Delta x_t - (W u_{t-\delta} + b) \bigr\|_2^2
    + \lambda_{\mathrm{R}}\, \|W\|_F^2,
\end{equation}
where $\mathcal{T}_\delta$ denotes timesteps for which $u_{t-\delta}$ is
available. The delay $\delta$ is selected from $\{0,1,\ldots,30\}$ by
validation MSE.

\noindent\textbf{Family II: velocity-augmented affine map.}
To capture short-horizon tracking effects without an explicit delay, we augment
the input with a finite-difference command velocity:
\begin{equation}
  \dot{u}_t = \frac{u_{t+1} - u_{t-1}}{2\,\Delta t},
  \qquad
  \Delta\hat{x}_t =
  W
  \begin{bmatrix}
    u_t \\
    \dot{u}_t
  \end{bmatrix}
  + b .
\end{equation}
This model is evaluated when neighboring commands are available, as in
demonstration fitting and full-chunk mapping.

\noindent\textbf{Family III: ARX map with FIR input history.}
The most expressive family uses $K+1$ finite-impulse-response input taps and
optionally $L$ autoregressive output taps:
\begin{equation}
  \Delta\hat{x}_t
  =
  \sum_{k=0}^{K} W^{\mathrm{in}}_k\, u_{t-k}
  +
  \sum_{l=1}^{L} W^{\mathrm{ar}}_l\, \Delta\hat{x}_{t-l}
  + b .
\end{equation}
The FIR terms capture latency and finite-bandwidth controller responses without
requiring a single fixed delay, while the optional AR terms capture low-order
tracking lag. We sweep
$(K,L)\in\{1,2,3,5,8\}\times\{0,1,2\}$ and select AR variants only when they
improve held-out rollout error.

\subsection{Family Selection}
\label{app:mapping-selection}

For each model family, we sweep its hyperparameters: the delay $\delta$ for
Family~I, the ridge weight
$\lambda_{\mathrm{R}} \in \{0,10^{-6},10^{-4},10^{-3},10^{-2},10^{-1},1,10\}$,
and the FIR/AR lag counts for Family~III. We select the configuration with the
lowest validation RMSE on a held-out 20\% split of the demonstration data, and
then refit the selected model on the full demonstration set. The sensitivity
sweep in Fig.~\ref{fig:mapping-sweep} shows that validation error saturates
after approximately five training demonstrations; therefore, the
50-demonstration fits used in production are not data-limited.
\begin{figure}[H]
  \centering
  \includegraphics[width=.99\linewidth]{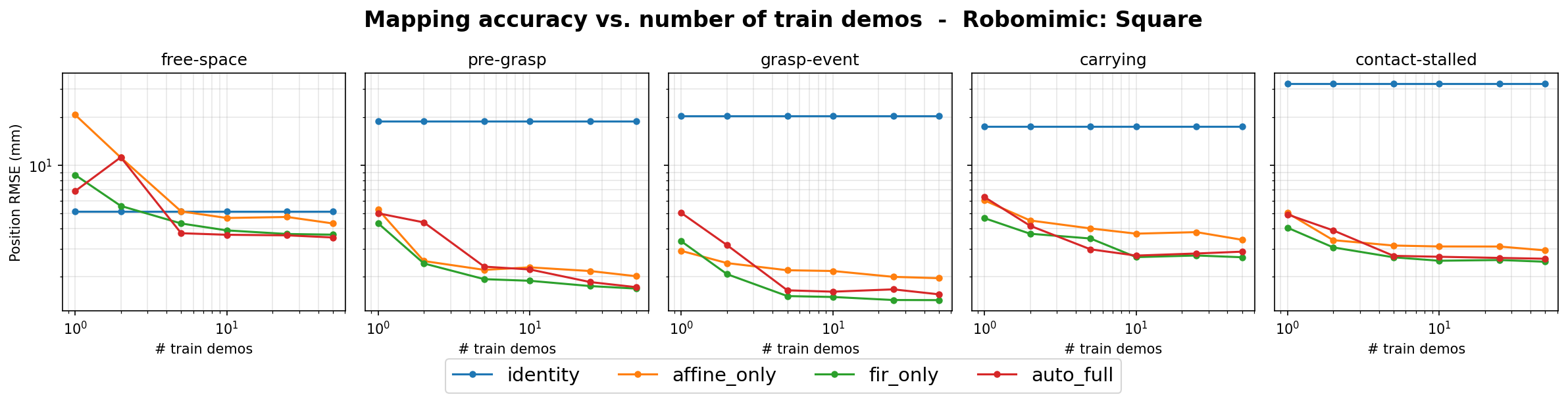}
  \caption{
Sensitivity of controller-aware mapping accuracy to the number of training
demonstrations. Validation RMSE decreases rapidly and saturates after
approximately five demonstrations, indicating that the 50-demonstration fits
used in production are not data-limited.
}
  \label{fig:mapping-sweep}
\end{figure}

\subsection{Stage 3: Trajectory Accumulator}
\label{app:mapping-accumulator}

The mapped per-step deltas $\Delta\hat{x}_{t:t+T_H-1}$ are integrated to
produce an end-effector trajectory that begins at the \emph{current measured
pose} $x_t=(p_t,q_t)$ and extends $T_H$ steps into the future:
\begin{align}
  \hat{p}_{t+k}
  &= p_t + \sum_{i=0}^{k-1} \Delta\hat{p}_{t+i},
  &\quad k = 1,\ldots,T_H \\
  \hat{q}_{t+k}
  &= \mathrm{Exp}(\hat{\omega}_{t+k-1}) \otimes \hat{q}_{t+k-1},
  \qquad \hat{q}_t = q_t,
  &\quad k = 1,\ldots,T_H
\end{align}
where $\Delta\hat{x}_{t+i}=(\Delta\hat{p}_{t+i},\hat{\omega}_{t+i})$,
$\otimes$ is quaternion multiplication, and $\mathrm{Exp}(\cdot)$ maps an
axis-angle vector to a unit quaternion. The left composition follows the
controller/world-frame convention after the Stage~1 frame correction. Anchoring
the accumulator to the live pose $x_t$ at every chunk refresh prevents mapping
errors from accumulating beyond one chunk horizon $T_H\le T_a$, providing
implicit closed-loop correction even though the mapping is open-loop within a
chunk.

\subsection{Implementation Notes}
\label{app:mapping-impl}

\noindent\textbf{Cross-chunk priming for FIR.}
For $K>0$, the lagged inputs $u_{t-k}$ at the start of a fresh chunk would be
zero under naive zero-padding, injecting an artificial step transient into the
prediction. We instead prime the FIR taps with the last $K$ actions of the
preceding chunk, preserving temporal continuity at chunk boundaries.
AR-augmented variants are not primed because their initial feedback terms would
come from the model's own previous predictions, which can amplify the rollout
instability discussed above.

\noindent\textbf{Single-step batched mode.}
The mapping is also queried per action inside the offline critic update to
compute the look-ahead reward $\tilde r^{\,\mathrm{TIGER}}$
(Eq.~\eqref{eq:tilde_r_igrl}). In this mode there is no temporal context, so we
evaluate only the lag-0 input block of the selected linear map and ignore FIR/AR
history terms. This approximation is used only to rank candidate actions inside
the conservative regularizer, not to generate TD targets or multi-step
rollouts.

\noindent\textbf{Per-robot vs. per-task fit.}
Because the mapping captures controller-induced action-to-motion mismatch rather
than task-specific behavior, one fit is reused across tasks when the robot,
controller mode, and gains are unchanged. We refit when changing the controller
configuration, action parameterization, or robot embodiment.

\noindent\textbf{Compute cost.}
Fitting the full family and hyperparameter sweep on approximately 50
demonstrations takes less than one second on a single CPU core. The selected
linear model adds at most $50\,\mu\mathrm{s}$ per chunk at inference, requiring
one matrix-vector product per predicted step, and introduces no measurable
overhead in the RL training loop.
\section{Goal-Aware CQL Pretraining: Look-Ahead and Objective Derivation}
\label{app:goal_aware_cql}

This appendix provides (i) the construction of the one-step
look-ahead oracle reward using the controller-aware trajectory
mapping, (ii) a derivation of the goal-aware conservative
objective used for offline pretraining as a $Q$-modulated
soft-max regularizer, (iii) an analysis showing that the
proposed modification preserves the conservative bias of CQL
while modulating its strength based on task geometry, and
(iv)~a comparison to two alternative aggregator forms we
considered but did not adopt.

\subsection{Controller-Aware One-Step Look-Ahead Signal}
\label{app:lookahead_signal}

Let $\mathcal{D}=\{(s_t,a_t,r_t,s_{t+1})\}$ denote the offline
dataset collected from demonstrations. Following
Sec.~\ref{subsec:imitation_guided_reward}, we define an IL-guided
navigation goal $x^g_t$ and augment the state as
$s'_t=(s_t,x^g_t)$. In our setting, $s_t$ contains high-dimensional
observations such as images, while the environment transition dynamics are
unknown. We therefore cannot evaluate the true next observation $s_{t+1}$
for arbitrary candidate actions.

Instead, we evaluate candidate actions through the same controller-aware
single-step surrogate used by the trajectory mapping. Given a candidate RL
action $a$ in the normalized controller action space and the current
proprioceptive end-effector pose $x_t$, we compute
\begin{equation}
  \hat{x}_{t+1}
  =
  F_{\mathrm{map}}^{\mathrm{1step}}(x_t,a),
  \label{eq:ctrl_map_appendix}
\end{equation}
where $F_{\mathrm{map}}^{\mathrm{1step}}$ denotes one-step evaluation of
the controller-aware mapping from
Sec.~\ref{subsec:controller_aware_trajectory_mapping}, including the known
controller transform, the selected residual model, and pose integration.
This produces an execution-aligned prediction of the next end-effector pose
for the candidate action, without requiring the true next observation.

This gives the one-step look-ahead alignment signal
\begin{equation}
  \tilde r^{\mathrm{TIGER}}_{t}(s'_t,a)
  =
  m_{t+1}\!\left(\hat{x}_{t+1},\, x^g_t\right),
  \label{eq:lookahead_reward_appendix}
\end{equation}
which can be computed for any dataset state $s_t\!\sim\!\mathcal{D}$ and
candidate action $a$ without observing the true next image. Since $m$ is a
product of two exponential pose-alignment factors,
$\tilde r^{\mathrm{TIGER}}_{t}\in[0,1]$
(see Sec.~\ref{subsec:imitation_guided_reward}).

\subsection{Background: CQL Conservative Regularizer}
\label{app:cql_background}

Conservative Q-Learning (CQL) augments the standard TD objective
with a conservative regularizer that suppresses $Q$-values on
actions sampled from a proposal distribution while preserving
values on dataset actions. Two functionally related forms appear
in the CQL literature: a \emph{linear} (Monte~Carlo) estimator
\begin{equation}
  \mathcal{R}_{\text{CQL}}^{\text{lin}}(\theta)
  =
  \mathbb{E}_{s\sim\mathcal{D},\, a\sim \mu(\cdot|s)}
  \!\left[ Q_\theta(s,a)\right]
  -
  \mathbb{E}_{(s,a)\sim\mathcal{D}}
  \!\left[ Q_\theta(s,a)\right],
  \label{eq:cql_reg_basic}
\end{equation}
and a \emph{soft-max} (log-sum-exp) estimator
\begin{equation}
\begin{aligned}
\mathcal{R}_{\text{CQL}}^{\text{sm}}(\theta)
&=
\mathbb{E}_{s\sim\mathcal{D}}
\Big[
\tau\log
\mathbb{E}_{a\sim\mu(\cdot\mid s)}
\big[\exp(Q_\theta(s,a)/\tau)\big]
\Big] \\
&\quad
-\mathbb{E}_{(s,a)\sim\mathcal{D}}
\big[Q_\theta(s,a)\big].
\end{aligned}
\label{eq:cql_reg_softmax}
\end{equation}
where $\mu(\cdot|s)$ is a proposal distribution over actions and
$\tau>0$ is a temperature. The soft-max form is better controlled 
in practice because the log-sum-exp smoothly concentrates on
high-valued candidate actions rather than applying uniform linear pressure to all sampled actions.

\subsection{Goal-Aware Conservatism via \texorpdfstring{$Q$}{Q}-Modulated Soft-max}
\label{app:derivation}

The CQL penalties in~\eqref{eq:cql_reg_basic}--\eqref{eq:cql_reg_softmax}
discourage high values on actions sampled from a proposal distribution
$\mu(\cdot|s)$, but they do not distinguish between actions that are
geometrically aligned with local task progress and actions that move away from
it. In robotic manipulation, this distinction is important: some candidate
actions may lie outside the demonstration action support while still moving the
end effector toward the IL-guided reference. We introduce task awareness by
defining a \emph{task-modulated value}
\begin{equation}
  \widetilde Q_\theta(s,a)
  \;\triangleq\;
  w(s,a)\,Q_\theta(s,a)
  \;=\;
  \bigl(1 - \alpha\,\tilde r^{\text{TIGER}}(s,a)\bigr)\,Q_\theta(s,a),
  \label{eq:tilde_Q_appendix}
\end{equation}
with weight
\begin{equation}
  w(s,a) = 1 - \alpha\, \tilde r^{\text{TIGER}}(s,a),
  \qquad \alpha\in[0,1],\;\;
  \tilde r^{\text{TIGER}}(s,a)\in[0,1]
  \;\Rightarrow\;
  w(s,a)\in[0,1].
  \label{eq:weight_linear}
\end{equation}
Replacing $Q_\theta$ by $\widetilde Q_\theta$
in~\eqref{eq:cql_reg_softmax}, while keeping the in-distribution
term anchored to the unmodulated value of dataset actions, gives the
goal-aware soft-max conservative regularizer used during TIGER pretraining:
\begin{equation}
\begin{aligned}
\mathcal{R}_{\text{GA-CQL}}(\theta)
&=
\mathbb{E}_{s\sim\mathcal{D}}
\Big[
\tau\log
\mathbb{E}_{a\sim\mu(\cdot\mid s)}
\big[\exp(\widetilde Q_\theta(s,a)/\tau)\big]
\Big] \\
&\quad
-\mathbb{E}_{(s,a)\sim\mathcal{D}}
\big[Q_\theta(s,a)\big].
\end{aligned}
\label{eq:gacql_softmax}
\end{equation}

The objective recovers standard CQL when $\alpha=0$. For $\alpha>0$,
actions with higher one-step TIGER progress receive a smaller effective value
inside the conservative soft-max. Thus, among the high-valued candidate actions
that dominate the CQL penalty, the regularizer applies less downward pressure
to actions predicted to make local task-space progress, while retaining the
standard dataset-action anchor through the unmodulated second term.

\subsection{Effect on Conservative Updates}
\label{app:pessimism_proof}

We analyze the gradient induced by~\eqref{eq:gacql_softmax}
to clarify how the proposed regularizer redistributes
CQL-style conservative pressure across candidate actions. This
analysis should be interpreted as an update-level explanation,
rather than a new lower-bound proof for CQL.

\noindent\textbf{Soft-max distribution.}
Define the modulated Boltzmann distribution induced by
$\widetilde Q_\theta$ at state $s$ over the proposal $\mu$:
\begin{equation}
  p_\tau(a\mid s)
  \;=\;
  \frac{\mu(a\mid s)\,\exp(\widetilde Q_\theta(s,a)/\tau)}
       {\mathbb{E}_{a'\sim\mu(\cdot|s)}
       \!\bigl[\exp(\widetilde Q_\theta(s,a')/\tau)\bigr]}.
  \label{eq:soft_max_density}
\end{equation}
This distribution identifies which sampled candidate actions
currently dominate the conservative soft-max.

\noindent\textbf{Gradient descent update.}
For this update-level analysis, we treat $Q(s,a)$ as
independently parameterized tabular values and hold the
proposal $\mu$ and look-ahead weight $w(s,a)$ fixed.
The resulting expressions describe direct updates to these
values, rather than the exact pointwise changes induced by
updating a shared neural critic.

At a fixed state $s$, differentiating the soft-max term gives
\begin{equation}
\nabla_{\!Q(s,a)}
\tau\log
\mathbb{E}_{a'\sim\mu(\cdot\mid s)}
\big[\exp(\widetilde Q(s,a')/\tau)\big]
=
w(s,a)\,p_\tau(a\mid s).
\end{equation}
Including the outer dataset-state expectation and the
dataset-action term, the conservative component induces
the update
\begin{equation}
\begin{aligned}
\Delta Q(s,a)
&\propto
-\lambda_{\mathrm{cql}}
\nabla_{\!Q(s,a)}\mathcal{R}_{\mathrm{GA\text{-}CQL}} \\
&=
\lambda_{\mathrm{cql}}
\Big(
d^{\mathcal{D}}(s,a)
-d^{\mathcal{D}}(s)\,w(s,a)p_\tau(a\mid s)
\Big),
\end{aligned}
\label{eq:grad_update_softmax}
\end{equation}
where $d^{\mathcal{D}}(s,a)$ is the empirical joint
state-action distribution and $d^{\mathcal{D}}(s)$ is
its state marginal. Thus, dataset-supported actions receive
upward pressure from the data term, while sampled actions
outside the dataset receive downward pressure through the
soft-max term. The latter is distributed according to
$\widetilde Q=wQ$ and scaled by the look-ahead weight.

When the proposal also depends on the dataset next state,
as in Eq.~\eqref{eq:pretraining_loss}, the corresponding
conditional soft-max gradients are additionally averaged
over dataset next states.

\noindent\textbf{Conservative pressure.}
For a dataset state with $d^{\mathcal{D}}(s)>0$, define
$d^{\mathcal{D}}(a\mid s)
=d^{\mathcal{D}}(s,a)/d^{\mathcal{D}}(s)$.
For dataset-supported actions, the data term counteracts
conservative suppression. For actions with
$d^{\mathcal{D}}(a\mid s)=0$, the conservative update is
proportional to
\begin{equation}
-\lambda_{\mathrm{cql}}\,
d^{\mathcal{D}}(s)\,w(s,a)p_\tau(a\mid s),
\end{equation}
which is nonpositive and suppresses unsupported actions
with positive look-ahead weight and soft-max mass.
The strength of this suppression depends on predicted
task-space progress.

\noindent\textbf{Goal-aware modulation.}
The look-ahead signal affects the conservative update through
two mechanisms. First, it directly scales the gradient magnitude
by $w(s,a)$ in~\eqref{eq:grad_update_softmax}. Second, it
changes the soft-max distribution itself because
$p_\tau(a\mid s)$ is computed from $\widetilde Q_\theta=wQ_\theta$
rather than from $Q_\theta$.

For misaligned actions, the look-ahead reward is small and
$w(s,a)\approx 1$, so the update approaches the standard CQL
penalty on high-valued unsupported actions. For task-aligned
actions, the look-ahead reward is larger and $w(s,a)<1$,
reducing the direct conservative gradient. Moreover, for the
high-valued positive candidates that typically dominate the
soft-max, replacing $Q$ by $wQ$ lowers their effective score
and further reduces their contribution to the conservative
penalty. In this sense, GA-CQL relaxes pessimism along
directions predicted to make local task-space progress, while
retaining strong CQL-style suppression for misaligned
candidate actions.

\subsection{Alternative Aggregator Forms Considered}
\label{app:alternative_forms}

We considered two alternative ways of combining the look-ahead
weight with the CQL conservative term during development. Both
are mathematically natural extensions of~\eqref{eq:cql_reg_basic}
or~\eqref{eq:cql_reg_softmax}; we report them here for
completeness and to motivate the choice
of~\eqref{eq:gacql_softmax}.

\noindent\textbf{(i) Linear Surrogate.}
Multiplying the linear estimator~\eqref{eq:cql_reg_basic} by $w$
yields
\begin{equation}
\begin{aligned}
\mathcal{R}_{\text{GA-CQL}}^{\text{lin}}(\theta)
&=
\mathbb{E}_{s\sim\mathcal{D},\, a\sim \mu(\cdot|s)}
\!\bigl[w(s,a)\,Q_\theta(s,a)\bigr] \\
&\quad-
\mathbb{E}_{(s,a)\sim\mathcal{D}}
\!\bigl[Q_\theta(s,a)\bigr].
\end{aligned}
\label{eq:gacql_linear_alt}
\end{equation}
This estimator admits a clean importance-reweighting
interpretation. Define
\begin{equation}
  Z(s) = \mathbb{E}_{a\sim\mu(\cdot|s)}[w(s,a)],
  \qquad
  \tilde\mu(a|s) = \frac{w(s,a)}{Z(s)}\,\mu(a|s),
  \label{eq:reweighted_dist}
\end{equation}
under which (assuming $w(s,a)\!\ge\!\varepsilon\!>\!0$ so that
$Z(s)\!>\!0$) one has
\begin{equation}
  \mathbb{E}_{a\sim \mu}[w(s,a)\,Q_\theta(s,a)]
  \;=\;
  Z(s)\,\mathbb{E}_{a\sim \tilde\mu}[Q_\theta(s,a)],
  \label{eq:reweighted_expectation}
\end{equation}
i.e., the linear form is equivalent to standard linear-CQL on the
reweighted proposal $\tilde\mu$ up to the state-dependent scaling
$Z(s)$. Empirically, however,
\eqref{eq:gacql_linear_alt} has no built-in bound on the
conservative pressure: when OOD $Q$ values are unconstrained the
term grows monotonically negative and the critic saturates within
$\sim\!15$ pretraining epochs in our Robomimic experiments
(Sec.~\ref{sec:Experiments}). Stabilizing it requires an
external floor on OOD $Q$, e.g.\ Cal-QL~\cite{calql}, which then
re-introduces the value oscillation that motivates GA-CQL in the
first place.

\noindent\textbf{(ii) Density-Reweighted Boltzmann.}
Substituting the reweighted proposal $\tilde\mu$
of~\eqref{eq:reweighted_dist} directly into the soft-max
form~\eqref{eq:cql_reg_softmax} gives
\begin{equation}
\begin{aligned}
\mathcal{R}_{\text{GA-CQL}}^{\text{Boltz}}(\theta)
&=
\mathbb{E}_{s\sim\mathcal{D}}
\Big[
\tau\log
\mathbb{E}_{a\sim\tilde\mu(\cdot\mid s)}
\big[\exp(Q_\theta(s,a)/\tau)\big]
\Big] \\
&\quad
-\mathbb{E}_{(s,a)\sim\mathcal{D}}
\big[Q_\theta(s,a)\big].
\end{aligned}
\label{eq:gacql_boltzmann_alt}
\end{equation}
with normalizer $Z(s)$ absorbed into $\tilde\mu$. This form
recovers the standard CQL fixed point under the reweighted
proposal but adds a per-batch self-normalization
($\tau\log Z(s)$) and importance weighting that we did not find
empirically necessary; the proposal density is also more
sensitive to outliers in $\tilde r^{\text{TIGER}}$ since
$\tilde\mu$ collapses onto the highest-weight candidate when
$w$ has heavy tails.

\noindent\textbf{Adopted form.}
The soft-max-over-$\widetilde Q$ regularizer
in~\eqref{eq:gacql_softmax} combines the boundedness
of~\eqref{eq:cql_reg_softmax} with the task-awareness
of~\eqref{eq:gacql_linear_alt}: it (i)~is bounded above for any
bounded $Q_\theta$; (ii)~recovers standard CQL exactly when the
TIGER signal is uninformative ($w\equiv 1$); (iii)~leaves the TD target unchanged
because $\widetilde Q_\theta$ enters only the
conservative regularizer; and (iv)~does not require Cal-QL
flooring or per-batch normalization. We adopt it as the
production form throughout the paper.

\subsection{Full Pretraining Loss}
\label{app:full_loss}

In practice, we combine the goal-aware conservative regularizer
with the standard TD loss:
\begin{equation}
  \mathcal{L}_{\text{pre}}(\theta)
  =
  \mathcal{L}_{\text{TD}}(\theta)
  +
  \lambda_{\text{cql}}\,
  \mathcal{R}_{\text{GA-CQL}}(\theta),
  \label{eq:full_pretrain_loss}
\end{equation}
with $\mathcal{R}_{\text{GA-CQL}}$ the soft-max regularizer
of~\eqref{eq:gacql_softmax}.

\noindent\textbf{Practical Notes.}
The look-ahead signal $\tilde r^{\text{TIGER}}$ is computed purely
in end-effector space and is excluded from the TD target; it is
used only to shape the conservative regularizer. The pretraining
stage is fully offline and requires no additional data beyond
$\mathcal{D}$. Cal-QL's~\cite{calql} $\max(Q,R^{\text{MC}})$
floor on OOD candidates is \emph{not} applied within the GA-CQL
soft-max: the log-sum-exp aggregator already bounds conservative
pressure, and stacking the floor on top destabilizes
$Q$-learning empirically. Cal-QL is therefore retained only in
the two pure-CalQL baselines reported
in Sec.~\ref{sec:Experiments}.

\section{Hyperparameters}
\label{app:hp}
Tables~\ref{tab:core_hparams}-\ref{tab:flow_matching_hp} define the hyper-parameters used in our experiments.

\begin{table}[t]
  \centering
  \renewcommand{\arraystretch}{1.1}
  \caption{Core algorithm hyper-parameters.}
  \label{tab:core_hparams}
  \begin{tabular}{@{}lll@{}}
    \toprule
    \textbf{Symbol} & \textbf{Description} & \textbf{Value} \\
    \midrule
    $\gamma$        & Discount factor              & 0.99      \\
    $\sigma$        & Exploration Std              & 0.1       \\
    $U$             & Update frequency             & 2         \\
    $c$             & Noise clip                   & 0.3       \\
    $LR$            & Adam learning rate           & \num{1e-4}\\
    $N_\text{batch}$& Mini-batch size              & 256       \\
    ---             & Actor : critic update ratio  & 1 : 1     \\
    ---             & Actor dropout                & 0.5       \\
    ---             & Actor MLP layers             & 2         \\
    ---             & Actor MLP hidden dim         & 1024      \\
    ---             & Actor layer norm             & True      \\
    $E$             & Q-ensemble size              & 2         \\
    ---             & Critic MLP layers            & 2         \\
    ---             & Critic MLP hidden dim        & 1024      \\
    ---             & Critic layer norm            & True      \\
    ---             & nstep                        & 3         \\
    \bottomrule
  \end{tabular}
\end{table}

\begin{table}[t]
  \centering
  \caption{TIGER common hyper-parameters.}
  \label{tab:env_hparams}
  \renewcommand{\arraystretch}{1.1}
  \begin{tabular}{@{}lll@{}}
    \toprule
    \textbf{Parameter} & \textbf{Description}                 & \textbf{Value} \\
    \midrule
    $\lambda_p$ & position distance scaling factor (eq.~\eqref{eq:pose_alignment})           & 5.0  \\
    $\lambda_q$ & rotation distance scaling factor (eq.~\eqref{eq:pose_alignment})           & 10.0 \\
    $\alpha$ & imitation guidance strength (eq.~\eqref{eq:task_aware_weighting}) & 1.0 \\
    $\lambda_\text{CQL}$ & CQL pretraining loss scale (eq.~\eqref{eq:pretraining_loss}) & 1.0 \\
    \bottomrule
  \end{tabular}
\end{table}

\begin{table}[!t]
  \centering
  \caption{TIGER environment-specific hyper-parameters.}
  \label{tab:hparams_tasks}
  \renewcommand{\arraystretch}{1.1}
  \setlength{\tabcolsep}{3pt}
  \begin{tabular}{@{}lccccc@{}}
    \toprule
    \textbf{Parameter} & \textit{Real Franka} & \textit{MetaWorld} & \textit{Robomimic} & \textit{Humanoid}\\
    \midrule
    $\lambda_\textit{TIGER}$ & 1.0  & 2.0  & 3.0 & 1.0  \\
    $\lambda_\textit{env}$  & 10.0 & 10.0  & 30.0   & 20   \\
    \bottomrule
  \end{tabular}
\end{table}

\begin{table}[!t]
  \centering
  \caption{Flow-matching imitation-policy hyper-parameters.}
  \label{tab:flow_matching_hp}
  \renewcommand{\arraystretch}{1.1}
  \begin{tabular}{@{}lll@{}}
    \toprule
    \textbf{Parameter} & \textbf{Description} & \textbf{Value} \\
    \midrule
    Backbone & Visual encoder backbone & ResNet-18 \\
    BatchNorm & Backbone normalization & FrozenBatchNorm2d \\
    Hidden dim & Per-camera feature dimension & 256 \\
    Fused dim & Conditioning feature dimension & 128 \\
    $T_a$ & Predicted action chunk length & 20 \\
    Time dim & Flow-time embedding dimension & 64 \\
    UNet base channels & Base channel width & 128 \\
    UNet channel multipliers & Resolution-level widths & $(1,2,2)$ \\
    Residual blocks & Blocks per resolution level & 1 \\
    Conditioning & Time and observation conditioning & FiLM \\
    $t$ sampling & Flow-time sampling during training & $\mathcal{U}(0,1)$ \\
    Initial noise & Inference initialization & $\epsilon\sim\mathcal{N}(0,I)$ \\
    Flow steps & Euler integration steps & 10 \\
    $T_H$ & Executed actions per query / replanning horizon & 10 \\
    \bottomrule
  \end{tabular}
\end{table}

\section{Network Architectures and Baseline Implementation Details}
\label{app:network_architectures}

This appendix describes the network architectures and baseline implementation
choices used in our experiments. TIGER, IBRL, and RLPD use the same
IBRL-style actor--critic architecture~\cite{ibrl}, consisting of a lightweight
ViT image encoder followed by actor and critic heads. TIGER differs from these
baselines only through the augmented target-pose observation, the
imitation-guided reward, and the offline value-pretraining objective. For
external baselines, we follow the original architectures and hyperparameters
whenever possible, and modify only the task-interface parameters required to
match our observation space, action space, controller frequency, and
demonstration format.

\subsection{IBRL-Style Networks (used by TIGER / IBRL / RLPD)}
\label{app:ibrl_networks}

TIGER, IBRL, and RLPD use the same IBRL-style actor--critic architecture. The
network consists of a lightweight ViT encoder for $96\times96$ RGB observations,
followed by separate actor and critic heads. The critic uses the token-wise
fusion architecture from IBRL: each visual channel is combined with
proprioception and action before aggregation, allowing the Q-function to
condition visual features on the candidate action. The actor instead flattens
the visual tokens, compresses them to a compact visual feature, concatenates
proprioception, and outputs a bounded $7$-D action. The layer-level specification is given below.

{\footnotesize
\begin{verbatim}
Visual encoder:
  input:
    RGB image, shape (3, 96, 96)

  patch embedding:
    Conv2D(3 -> 128, kernel=8, stride=4)
    ReLU
    Conv2D(128 -> 128, kernel=3, stride=2)

  transformer block:
    LayerNorm
    MultiHeadAttention(dim=128)
    LayerNorm
    Linear(128 -> 512)
    Linear(512 -> 128)

  output:
    121 visual tokens, each 128-D
    shape = (121, 128)

Critic Q(s, a):
  inputs:
    visual tokens, shape (121, 128)
    proprioception, dimension P
    action, dimension A

  token/action fusion:
    transpose tokens: (121, 128) -> (128, 121)

    for each visual channel:
      concatenate [token vector, proprioception P, action A]
      Linear(121 + P + A -> 1024)

    element-wise product with learned channel weights,
    shape (128, 1024)

    sum over visual channels

  output:
    1024-D critic feature

  Q-head:
    concatenate [critic feature, action A]
    Linear(1024 + A -> 1024)
    LayerNorm
    ReLU
    Linear(1024 -> 1024)
    LayerNorm
    ReLU
    Linear(1024 -> 1)

  output:
    scalar Q-value

Actor pi(s):
  inputs:
    visual tokens, shape (121, 128)
    proprioception, dimension P

  visual feature:
    flatten tokens: 121 * 128 = 15488
    Linear(15488 -> 128)
    LayerNorm
    Dropout(0.5)
    ReLU

  policy head:
    concatenate [visual feature, proprioception P]
    Linear(128 + P -> 1024)
    LayerNorm
    Dropout(0.5)
    ReLU
    Linear(1024 -> 1024)
    LayerNorm
    Dropout(0.5)
    ReLU
    Linear(1024 -> 7)
    Tanh

  output:
    7-D action
\end{verbatim}
}
\subsection{External Baselines}
\label{app:external_baselines}

For LaNE, ResFiT, and DSRL, we use the original implementations and default
hyperparameters unless otherwise stated. When a baseline admits multiple
reasonable settings for our action-chunking setup, we evaluate the alternatives
and report the stronger-performing configuration. This gives each baseline the
benefit of a task-compatible implementation while preserving the original
method design.

\noindent\textbf{LaNE.}
For LaNE~\cite{LaNE}, we use the default architecture, optimization settings,
and hyperparameters from the original implementation. We only adapt the
environment interface to match our observation and action spaces. In
particular, LaNE is trained with the same action
parameterization and environment-step budget.

The LaNE latent dynamics model uses latent dimension $16$ and action dimension
$7$. The DINOv2-based encoder takes two $3\times112\times112$ random crops as
input and uses the \texttt{dinov2\_vits14\_reg} backbone. The two DINOv2
embeddings are concatenated and passed through three $512$-D ReLU MLP layers,
followed by a linear layer producing $32$ outputs, corresponding to a
$16$-D mean and $16$-D log standard deviation. The decoder mirrors this MLP and
predicts two $384$-D DINOv2 embeddings. For the non-DINOv2 image encoder, LaNE
uses stacked convolutional layers followed by two $128$-D ReLU MLP layers and
a $32$-D output. The locally linear dynamics model predicts
$u\in\mathbb{R}^{16}$, $v\in\mathbb{R}^{16}$,
$B\in\mathbb{R}^{16\times7}$, and offset $c\in\mathbb{R}^{16}$. The RL
encoder maps $6\times112\times112$ image inputs to a $32$-D feature, followed
by actor and critic heads with two $1024$-D hidden layers.

\noindent\textbf{ResFiT.}
For ResFiT~\cite{resfit}, we use the default architecture and training
hyperparameters from the original implementation, adapting only the task
interface to our observation and action spaces. In simulation, we use the
standard ResFiT warmup protocol, including $10$k environment warmup steps and
$10$k critic warmup updates before online actor--critic training.

In simulation, we evaluate ResFiT with the default actor learning rate
$10^{-6}$ and with a larger actor learning rate $10^{-5}$, and report the
better-performing configuration for each task. We include this sweep because
the default actor update was overly conservative in several of our settings.
We also observed that ResFiT is sensitive to random seed in several
sparse-reward tasks. This effect was most pronounced in the humanoid
shape-sorting task, where some seeds failed to discover successful behavior
while others reached high final success. We therefore report ResFiT using the
stronger of the two actor learning rates evaluated above, while retaining the
same number of seeds and evaluation protocol used for the other methods.

For real-robot experiments, we do not use environment warmup steps, so that all
methods are compared under the same online interaction budget. We use $1$k
critic warmup updates before actor--critic training and set the actor learning
rate to $10^{-5}$, which was the stronger and more stable configuration in
preliminary runs.

\noindent\textbf{DSRL.}
For DSRL~\cite{dsrl}, we use DSRL-SAC as the primary implementation. DSRL
fine-tunes a frozen behavior-cloning policy by learning an RL policy in the
latent noise space of the BC model, rather than directly learning residuals in
the executed action space. For chunked policies, the full latent-noise action
has dimension $C \times d$, where $C$ is the action chunk size and $d$ is the
per-step noise dimension. Training an actor and critic over this full
high-dimensional noise vector can be difficult. Following the DSRL paper, we
first evaluated the recommended single-step noise parameterization, where SAC
learns a policy over $w_{\mathrm{single}}\in\mathbb{R}^{d}$ and repeats this
same noise vector across the chunk dimension before querying the frozen BC
policy.

In our setting, this repeated-noise parameterization performed poorly with the
long-horizon BC policy used by TIGER, which predicts chunks with horizon
$H=20$ and action step $A_s=10$. We therefore trained a shorter-horizon BC
policy on the same data with $H=4$ and $A_s=4$ and re-ran DSRL under this setting. The shorter
chunk reduces the dimensionality of the latent-noise control problem and allows
DSRL to predict distinct per-step noise variables across the four-step chunk,
rather than repeating a single noise vector over a long horizon. In simulation,
we evaluate both DSRL-SAC and DSRL-NA with this $H=4$, $A_s=4$ BC policy and
report the better-performing variant. For real-robot experiments, we also use
the $H=4$, $A_s=4$ BC policy, since this configuration performed better than
the longer-horizon $H=20$, $A_s=10$ policy for DSRL in our setup.

\paragraph{Flow-matching imitation policy.}
Our imitation policy uses an ACT-style~\cite{act} visual encoder followed by a
conditional 1D UNet flow-matching action head. Each camera image is processed independently by an
ImageNet-pretrained ResNet-18 backbone with frozen batch normalization. The
resulting $512$-channel feature map is projected to $256$ channels with a
$1\times1$ convolution and spatially average-pooled to a $256$-D vector. Robot
proprioception is projected to the same dimension with a linear layer. The
camera features and proprioceptive feature are concatenated and passed through
a small MLP with LayerNorm, ReLU, and dropout to produce a $128$-D conditioning
vector.

The action head predicts action chunks of length $H$. In our experiments,
$H=20$ for all tasks, and we execute the first $10$ actions from each
chunk before replanning. The head is a conditional 1D UNet operating along the
chunk horizon. It uses base width $128$, three resolution levels with channel
multipliers $(1,2,2)$, and FiLM conditioning from the concatenated visual
conditioning vector and diffusion-time embedding. Each residual block contains
GroupNorm, SiLU activations, Conv1d layers, dropout, and a residual skip
connection. The final layer projects the UNet output back to the action
dimension.

We train the policy with conditional flow matching. Given an expert action
chunk $x_{\mathrm{start}}$ and Gaussian noise $\epsilon$, we sample
$t\sim\mathcal{U}(0,1)$ and form
\[
x_t = t x_{\mathrm{start}} + (1-t)\epsilon .
\]
The network predicts the target velocity
\[
v^\star = x_{\mathrm{start}} - \epsilon
\]
using an MSE loss. At inference time, we initialize the chunk from Gaussian
noise and integrate the learned flow using $10$ Euler steps on a linear
$t\in[0,1]$ schedule.
\fi
\end{document}